\documentclass[sigconf,authorversion,nonacm]{acmart}
\AtBeginDocument{%
  }

\usepackage{booktabs}
\usepackage{multirow}
\usepackage{graphicx}
\usepackage{subcaption}
\usepackage{amsmath}
\usepackage{mathtools}
\usepackage{algorithm}
\usepackage{algpseudocode}
\usepackage{xcolor}
\usepackage{hyperref}
\usepackage{enumitem}

\usepackage{xcolor}

\definecolor{my-dark-yellow}{RGB}{249, 189, 0}
\definecolor{my-dark-blue}{RGB}{31, 138, 204}

\newcommand{\doPair}{\textsc{DO} pair}
\newcommand{\CR}{\textsc{TC}}

\newcommand{\CBA}{\textsc{PCBA}}
\newcommand{\PCAA}{\textsc{PCAA}}
\newcommand{\DCAA}{\textsc{DCAA}}
\newcommand{\ETA}{\textsc{ETA}}

\newcommand{\PWT}{\textsc{PWT}}

\newcommand{\method}{\textsc{EXHOLD}}

\usepackage{xcolor}
\definecolor{pos}{RGB}{0,120,60}    
\definecolor{neg}{RGB}{180,40,40}   
\definecolor{neu}{gray}{0.35}

\newcommand{\posdelta}[1]{\textcolor{pos}{#1}}
\newcommand{\negdelta}[1]{\textcolor{neg}{#1}}

\newcommand{\hb}{\textcolor{neu}{(\,$\uparrow$\,)}}
\newcommand{\lb}{\textcolor{neu}{(\,$\downarrow$\,)}}

\newcommand\blfootnote[1]{%
  \begingroup
  \renewcommand\thefootnote{}\footnote{#1}%
  \addtocounter{footnote}{-1}%
  \endgroup
}

\begin{document}

\title[EXHOLD: Experience-Aware Ride-Hailing Hold Control]{EXHOLD: Experience-Aware Real-Time Hold Control for Large-Scale Ride-Hailing Matching at DiDi}

\author{Xu Liu}
\affiliation{%
  \institution{Didichuxing Co. Ltd}
  \city{Beijing}
  \state{}
  \country{China}}
  \email{leoliuxu@didiglobal.com}

\author{Kai Wan}
\affiliation{%
  \institution{Didichuxing Co. Ltd}
  \city{Beijing}
  \state{}
  \country{China}}
  \email{peterwan@didiglobal.com}

\author{Zihao Lu}
\affiliation{%
  \institution{Didichuxing Co. Ltd}
  \city{Beijing}
  \state{}
  \country{China}}
  \email{luzihao@didiglobal.com}

\renewcommand{\shortauthors}{Xu et al.}


\begin{abstract}
In large-scale ride-hailing matching systems, \emph{hold control} is a high-leverage mechanism for improving end-to-end passenger--driver experience: by selectively delaying certain driver--order pairs, the system can wait for better opportunities, reduce cancellations and excessive waiting, and mitigate wasted driver effort, ultimately increasing trip success.
In practice, many industrial hold strategies rely on heuristic thresholding over multiple predictive models, which can be brittle under non-stationary traffic conditions and hard to optimize for multiple, experience-oriented objectives.

We propose \method, a deployable two-stage framework that decouples \emph{experience-aware pair assessment} from \emph{hold-time execution}.
In Stage~I, we learn a decision model that assigns each driver--order pair to discrete and interpretable \emph{experience tiers}, optimizing a unified objective that aggregates multiple satisfaction-related signals across the matching funnel.
In Stage~II, we solve for a monotone hold-time schedule via constrained optimization over empirical quantiles, explicitly enforcing service guardrails that bound unnecessary holding of promising matches while maximizing overall experience improvement.

We evaluate \method\ through online randomized A/B experiments in DiDi’s production ride-hailing matching system in Brazil.
The results demonstrate consistent gains in both marketplace efficiency and passenger--driver experience: \method\ increases trip completion and driver income, while significantly reducing passenger cancellations before and after acceptance and improving key funnel efficiency signals such as faster call-to-acceptance.
We further conduct targeted ablations and behavioral drill-down analyses, showing that both stages of \method\ are essential to the observed gains, and that the policy makes calibrated decisions under spatiotemporal heterogeneity.
\method\ has been ramped up and is currently serving the Brazil market in production.

\end{abstract}





\maketitle
\blfootnote{Accepted at the 32nd ACM SIGKDD Conference on Knowledge Discovery and Data Mining (KDD 2026). This is an arXiv author version.}

\section{Introduction}
\label{sec:intro}

Large-scale ride-hailing platforms solve a continuous stream of matching decisions under strict latency, reliability, and service-quality constraints~\cite{xu2018large}.
For each incoming order, the system evaluates many candidate driver--order pairs (\doPair) and must decide not only \emph{who} to match, but also \emph{when} to match~\cite{wang2021secure, zhang2017taxi}.
A critical lever in this process is \emph{hold control}: selectively holding certain \doPair{} candidates for some system-decided time so the system can wait for better opportunities, reduce cancellations and excessive waiting, and mitigate wasted driver effort---ultimately improving end-to-end passenger--driver experience and trip success~\cite{tu2024towards, afeche2023ride}.
Hold control is powerful but delicate: an overly aggressive policy may suppress good matches and harm service levels of the platform, while an overly conservative policy yields little impact~\cite{xu2018large, afeche2023ride}.

\paragraph{Why hold control is challenging in production.}
Hold decisions are inherently multi-objective, sequential, and tightly coupled with the matching pipeline~\cite{supian2024ride}.
First, experience degradation manifests through \emph{heterogeneous funnel outcomes}: Passenger Cancellation Before driver Acceptance (\CBA), driver non-response, Passenger Cancellation After driver Acceptance (\PCAA), as well as Driver Cancellation After Acceptance (\DCAA), have distinct user harm and efficiency implications~\cite{xu2018large, wang2021secure}.
Second, hold is a delayed-action mechanism: holding a candidate changes future states such as passenger waiting, local supply--demand regimes, and the set of feasible matches the system sees later~\cite{afeche2023ride, barbour2019embracing}.
Besides, production systems require explicit \emph{service guardrails}: the system must bound unnecessary holding of promising matches and remain stable under non-stationary traffic conditions~\cite{yan2020dynamic, suhr2019two}.

\paragraph{Limitations of heuristic model-threshold strategies.}
Many industrial hold modules are implemented as heuristic thresholding over multiple predictive models, for example mapping each driver-order pair to a fixed or dynamic threshold and then comparing a predictor against that threshold to decide hold or release~\cite{chen2021short, wen2024survey, yatnalkar2019machine}.
While effective as a baseline, this paradigm faces three recurring limitations in practice:
(i) \emph{Error compounding} arises when multiple models are chained and small calibration shifts amplify through the decision rule~\cite{somalwar2025learning};
(ii) \emph{Weak multi-objective coordination} makes it difficult to simultaneously optimize multiple experience-related signals across the funnel (e.g., \CBA, \PCAA, \DCAA\ and passenger over-waiting) with hand-tuned threshold rules~\cite{naumov2023optimizing};
(iii) \emph{Limited temporal reasoning} means temporal effects are typically approximated by static calibrations, which can be brittle under traffic shifts.
As a result, these strategies can be hard to tune for experience improvements under spatiotemporal heterogeneity and may not provide explicit, auditable control over ``how much holding is acceptable'' under service constraints~\cite{lee2022impact, tirachini2020ride, yang2020phase}.

\begin{figure*}[t]
    \centering
    \includegraphics[width=1.00\linewidth]{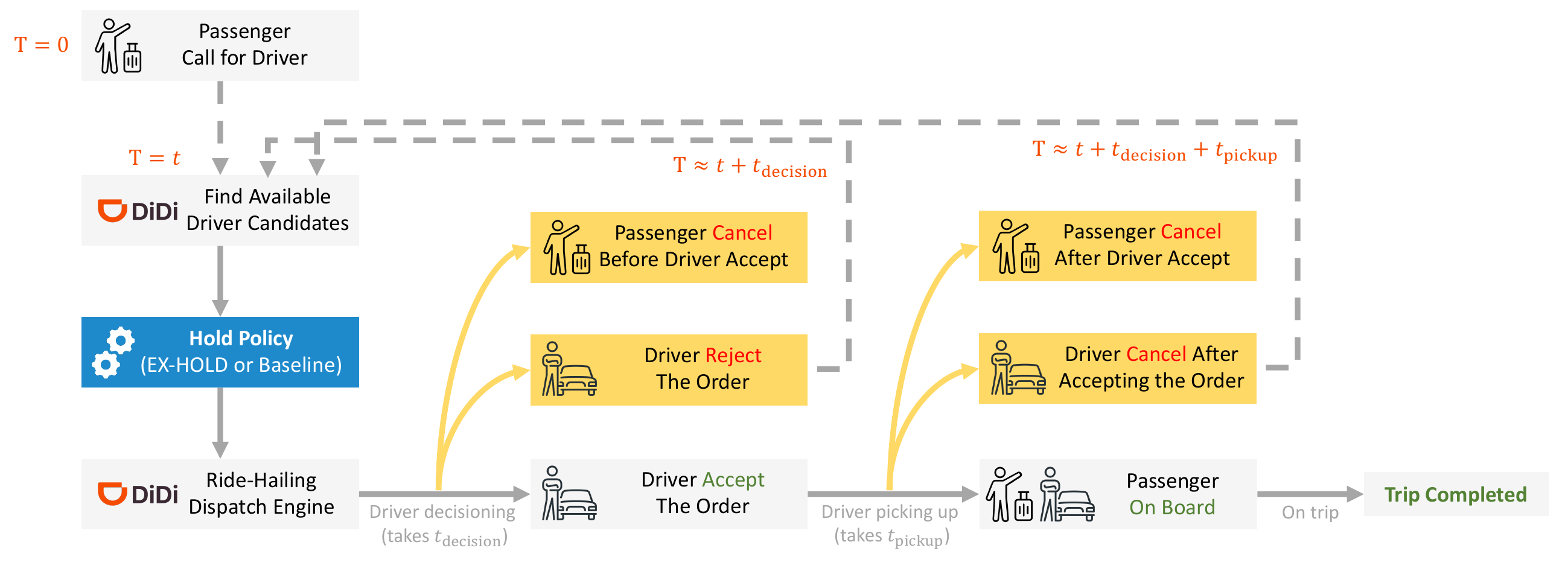}
    \caption{Schematic of heterogeneous experience feedback in ride-hailing passenger-driver matching. At timestep $t$, the system first identifies all available candidate drivers. A hold-decision module (highlighted in \textcolor{my-dark-blue}{blue}) then decides which drivers should be held for some time, while the remaining candidates are passed to the dispatch engine for broadcasting. During the driver’s decision-making and the subsequent pickup, various heterogeneous experience feedback may occur (highlighted in \textcolor{my-dark-yellow}{yellow}).}
    \label{fig:transition}
\end{figure*}

\paragraph{Our approach: decouple experience-aware assessment from hold-time execution.}
We propose \textbf{EX}perience-aware \textbf{HOLD} (\textbf{\method}), a deployable two-stage framework for hold control in industrial ride-hailing matching systems (Figure~\ref{fig:overview}).
The key idea of \method\ is to separate \emph{what to hold} from \emph{how long to hold} in a way that aligns with production constraints and experience-oriented objectives:
\begin{itemize}[leftmargin=*]
  \item \textbf{Stage I (Experience-aware pair assessment).}
  We learn a decision model that assigns each \doPair{} to a discrete, interpretable \emph{experience tier}.
  The tiering optimizes a unified objective that aggregates multiple passenger-driver satisfaction-related signals across the matching funnel (order completion, cancellations at different stages, and waiting-related indicators, shown in Figure~\ref{fig:transition}).
  To support robustness under non-stationary traffic, Stage~I introduces Transformer-based temporal representations~\cite{vaswani2017attention} and LinUCB~\cite{chu2011contextual} head that provides stable online decisions.
  \item \textbf{Stage II (Guardrail-constrained hold-time optimization).}
  Given the experience tiers from Stage~I, we compute a monotone hold-time schedule via constrained optimization over empirical quantiles~\cite{kotary2021end, liang2022survey}.
  This stage explicitly enforces service guardrails that bound unnecessary holding of promising matches and provides an controllable tool for conservativeness, enabling safe deployment at production platforms.
\end{itemize}

The decomposition design of \method\ is not merely an engineering convenience, but also a principled response to the difficulty of optimizing hold control strategies end-to-end under heterogeneous feedback, strict operational constraints, and the need for explicit guardrails in production~\cite{feng2021we, liu2025adaptive}.

\paragraph{Real-world deployment and key findings.}

We deploy \method\ in DiDi's production ride-hailing matching system in Brail and evaluate it through large-scale A/B experiments.
The results demonstrate that \method\ delivers consistent improvements on trip success and driver welfare while reducing experience-degrading cancellations and improving matching efficiency, indicating better end-to-end passenger--driver matching experience in real operations.
Beyond aggregate outcomes, we provide (i) funnel-level analyses that attribute the gains to improved conversion after exposure and fewer cancellations, and (ii) ablations and behavioral drill-downs that validate the role of multi-signal experience modeling and guardrail-constrained hold-time optimization.
After the A/B validation, \method\ has been ramped up and is now deployed at market scale, serving real traffic across the Brazil marketplace.

\paragraph{Contributions.}
We summarize our main contributions as follows:
\begin{itemize}[leftmargin=*]
  \item We formulate hold control in ride-hailing matching as a decision problem centered on passenger--driver experience, with explicit service guardrails to limit unnecessary holding.
  \item We propose \method, a practical two-stage framework that decouples experience-aware tier assignment from hold-time execution, combining representation learning, contextual bandits, and constrained optimization over empirical quantiles.
  
  

  \item We develop an experience-oriented objective and training pipeline that aggregates heterogeneous signals across the ride-hailing funnel. Online A/B experiments show improved trip success and driver welfare, reduced experience-degrading cancellations, and faster funnel efficiency; online ablations further confirm that both Stage I and Stage II are essential for consistent gains.

\end{itemize}

\section{Problem Formulation}
\label{sec:problem}

We formalize hold control in large-scale ride-hailing matching as a constrained decision problem with \emph{experience-oriented objectives} and \emph{service guardrails}.
Our formulation highlights industrial aspects that materially affect both optimization and evaluation: (i) the \doPair{} decision unit and its position in the matching pipeline, (ii) heterogeneous funnel outcomes that correspond to distinct passenger/driver experience costs, (iii) delayed effects through time and passenger--driver interactions, and (iv) explicit guardrails that bound unnecessary holding of promising matches.

\subsection{System Context and Decision Unit}
\label{sec:pf_context}

A production ride-hailing platform continuously evaluates candidate driver--order pairs for each incoming order.
We define the atomic \emph{decision unit} as a driver--order pair (\doPair), representing a specific order evaluated with a specific nearby driver at a particular time.
Hold control is applied to these candidates before they are executed by the downstream broadcast/assignment module, and thus can change both the set of candidates exposed to drivers and the composition of accepted and completed trips.

Specifically, at decision step $t$, the system observes a state vector $s_t \in \mathcal{S}$ for a \doPair{} that aggregates:
\begin{itemize}[leftmargin=*]
  \item \textbf{Order context:} current passenger waiting time and its evolution, OD attributes, fare/distance proxies, and other order features;
  \item \textbf{Driver context:} driver availability/status, recent activity patterns, historical response statistics, and other driver features;
  \item \textbf{Spatio-temporal feature:} time-of-day, local supply--demand conditions (e.g., nearby available drivers and orders), historical supply and demand, and other spatiotemporal features;
  \item \textbf{System feature:} estimated pickup distance and time, estimated trip time, and other system-augmented matching features.
\end{itemize}
These signals jointly characterize both \emph{immediate feasibility} (e.g., pickup distance/time) and \emph{experience sensitivity} (e.g., accumulated waiting and supply-demand stress).

\subsection{Experience Tiers and Hold-Time Execution}
\label{sec:pf_action}

A hold decision determines whether a \doPair{} should be released immediately to the downstream pipeline or temporarily delayed.
Directly optimizing a continuous hold duration online is difficult in production due to delayed credit assignment, unsafe exploration, and tight latency budgets. We therefore model the online decision as selecting a discrete \emph{experience tier}:
\begin{equation}
  a_t \in \mathcal{A} = \{0,1,\dots,K\},
  \label{eq:action_space}
\end{equation}
where larger $a_t$ indicates a less favorable pairing from an experience perspective and thus warrants stronger holding.
Each tier is subsequently mapped to a hold time $x_{a_t}$ computed offline by a constrained optimization procedure (Stage~II, Section~\ref{sec:stage2}).
Online execution applies the delayed-action policy:
\begin{equation}
  \text{hold duration} = x_{a_t}, \qquad x_0 = 0 \le x_1 \le \cdots \le x_K.
\end{equation}
This design keeps online inference fast and bounded while making the timing component explicitly controllable via guardrails.

\subsection{Heterogeneous Experience Costs for Hold}
\label{sec:pf_outcomes}

After a \doPair{} is released (immediately or after holding), the system evolves according to passenger and driver behaviors and downstream matching.
Each decision leads to one of several mutually exclusive outcomes: CBA, driver non-response, PCAA ir DCAA; these time-delayed outcomes are shown in Figure~\ref{fig:transition}.
These outcomes correspond to heterogeneous experience costs: \CBA\ is often tied to pre-accept friction and excessive waiting, \PCAA\ directly wastes driver effort and leads to passenger bad experience, and \DCAA\ can amplify inefficiency and subsequent cancellations.
Therefore, hold control must optimize \emph{multi-signal experience objectives} rather than a single proxy outcome.

\subsection{Hold Objective: Experience Utility Maximization Under Service Guardrails}
\label{sec:pf_objective}

Ideally, we seek a policy $\pi$ that maps \doPair{} states to experience tiers to maximize expected experience utility while satisfying explicit service guardrails.
Let $y_t \in \mathcal{Y}$ denote the realized outcome category and let $r_\text{oracle}(s_t,a_t,y_t)$ be the oracle reward that aggregates multiple experience-related outcomes, where $s_t$ is the real-time \doPair\ state (Section~\ref{sec:pf_context}), $a_t$ is the real-time hold action generated by the policy (whether to hold and how long to hold).
Then the constrained hold objective can be expressed as:
\begin{align}
  \max_{\pi} \quad &
  \mathbb{E}_{\pi}\!\left[\sum_{t} \gamma^t \, r_\text{oracle}(s_t,a_t,y_t)\right], \label{eq:pf_obj} \\
  \text{s.t.}\quad &
  \mathbb{P}_{\pi}\!\left(\text{unnecessary-hold} \mid y_t=\CR\right) \le \alpha, \label{eq:pf_constraint}
\end{align}
where $\gamma\in(0,1]$ is the discount factor and $\alpha$ is the tolerated rate of unnecessarily holding cases that would have been a complete trip.
Constraint~\eqref{eq:pf_constraint} captures an essential production guardrail: improving experience by reducing cancellations and excessive waiting should not come at the cost of suppressing too many promising matches, which would degrade service levels and harm passenger-driver experience. In practice, however, the oracle $r_\text{oracle}$ is typically unavailable at the decision time, as it depends on future outcomes. This necessitates the design of an effective proxy reward to realize experience-aware holding policies (Section~\ref{sec:reward_design}).

\section{Method}
\label{sec:method}

\begin{figure*}[t]
    \centering
    \includegraphics[width=0.9\linewidth]{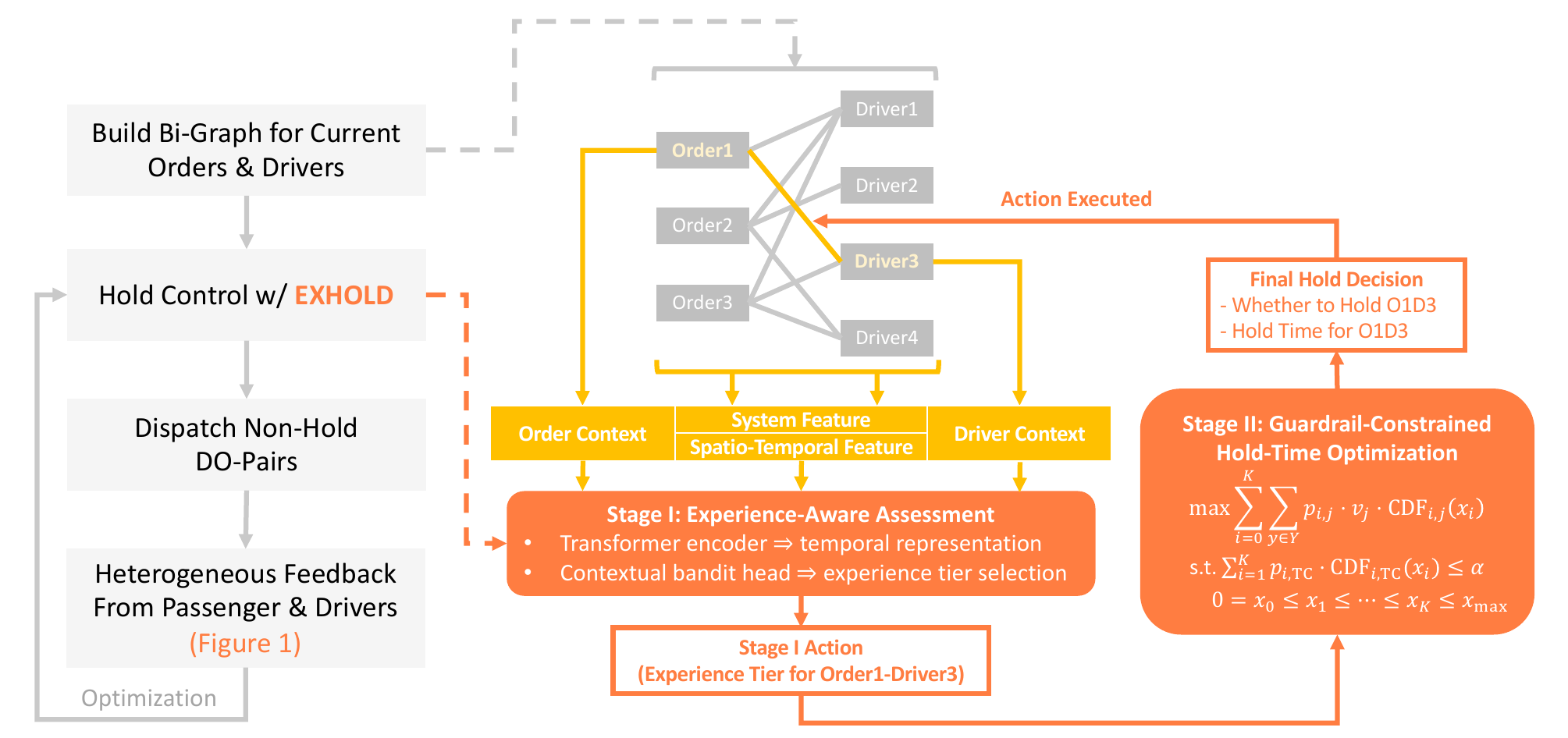}
    \caption{Overview of \textcolor{orange}{\method}. Stage~I produces discrete \emph{experience tiers} for each driver--order pair from rich context; Stage~II computes a monotone hold-time table under explicit service guardrails via constrained optimization.}
    \label{fig:overview}
\end{figure*}

We present \textbf{EX}perience-aware \textbf{HOLD} (\textbf{\method}), a deployable two-stage framework for experience-aware real-time hold control in large-scale ride-hailing matching.
The core design principle of \method\ is to decouple \emph{experience-aware pair assessment} (what to hold, Section~\ref{sec:stage1}) from \emph{hold-time execution} (how long to hold, Section~\ref{sec:stage2}) under explicit service guardrails.

\subsection{Method Overview}
\label{sec:overview}

At each decision step $t$, the system evaluates a candidate driver--order pair (\doPair) with state $s_t$ (Section~\ref{sec:pf_context}).
Stage~I assigns each \doPair{} to a discrete and interpretable experience tier $a_t \in \mathcal{A}=\{0,1,\dots,K\}$ with Transformer and contextual bandits, where larger $a_t$ indicates lower expected experience (e.g., higher passenger/driver cancellation propensity or worse waiting outcomes).
Stage~II maps each tier $i$ to a monotone hold-time table $\{x_i\}_{i=0}^{K}$ computed with constrained optimization.


\subsection{Stage I: Experience-Aware Assessment}
\label{sec:stage1}

Stage~I learns a policy $\pi_{\theta,\phi}$ that assigns each \doPair{} to an experience tier using
(i) a Transformer encoder that summarizes short-horizon temporal context and
(ii) a stable contextual bandit head that enables robust online decisions with controlled exploration.
This hybrid design balances representation power, stability under distribution shift, and production constraints.

\subsubsection{Sequential Context Construction}
\label{sec:context}

Hold decisions depend on time-evolving order state (e.g., accumulated waiting) and local regime changes (e.g., supply--demand shifts).
For each \doPair{} at time $t$, we construct a short history window $h_{\text{\tiny{DO}},t}$ composed of:
\begin{itemize}[leftmargin=*]
  \item \textbf{Order trace:} recent evolution of order-level state over the past $L$ steps (e.g., waiting progression and other state changes);
  \item \textbf{Local regime trace:} recent regime summaries (e.g., spatiotemporal buckets capturing supply--demand conditions);
  \item \textbf{Static attributes:} time-invariant or slowly varying attributes of the order, driver, and the DO pair.
\end{itemize}

\subsubsection{Transformer Encoder for Temporal Representation}
\label{sec:dt}

We encode $h_t$ using a Transformer encoder to obtain an embedding:
\begin{equation}
z_{\text{\tiny{DO}},t} = f_{\theta}(h_{\text{\tiny{DO}},t})\in\mathbb{R}^{p}.
\label{eq:embedding}
\end{equation}
The embedding is trained to summarize both instantaneous pair feasibility and short-horizon temporal patterns that correlate with experience degradation (e.g., cancellations and waiting outcomes).

Unlike purely supervised thresholding, Stage~I learns representations that are directly aligned with downstream hold control by optimizing an experience-oriented learning signal that aggregates multiple funnel outcomes.
Concretely, we attach a lightweight contextual bandit head on top of $z_{\text{\tiny{DO}},t}$ and train it with Huber loss in our implementation to predict a scalar utility reward; the exact construction of this reward is detailed in Section~\ref{sec:reward_design}, and the training details of the encoder are explained in Appendix~\ref{sec:appendix_alt_training}.

\subsubsection{Contextual Bandit Head for Experience Tier Decision}
\label{sec:linucb}

Given the \doPair{} embedding $z_{\text{\tiny{DO}},t}$ from the Transformer encoder at decision step $t$, we deploy a LinUCB contextual bandit head to select an experience tier with controlled exploration.
For each action $a\in\mathcal{A}$, the contextual bandit maintains a ridge-regression estimator and its design matrix computed from interaction data:
\begin{equation}
    A_a = \lambda_0 I + \sum_{\tau \in \mathcal{T}_a} z_{\text{\tiny{DO}},\tau}\, z_{\text{\tiny{DO}},\tau}^{\top}, \quad b_a = \sum_{\tau \in \mathcal{T}_a} z_{\text{\tiny{DO}},\tau}\, r_\tau,
    \label{eq:linucb_model}
\end{equation}
where $\mathcal{T}_a=\{\tau < t: a_\tau=a\}$ is the set of historical steps where tier $a$ was selected, $\lambda_0 > 0$ is the regularization coefficient, and $r_\tau$ is the experience-oriented reward defined in Section~\ref{sec:reward_design}, where $r_\tau$ is the realized experience reward logged for step $\tau$.
At runtime, the action is selected by an upper-confidence bound:
\begin{equation}
a_t = \arg\max_{a\in\mathcal{A}}
\left( z_{\text{\tiny{DO}},t}^{\top} A_a^{-1} b_a + \lambda \sqrt{ z_{\text{\tiny{DO}},t}^{\top} A_a^{-1} z_{\text{\tiny{DO}},t} } \right),
\label{eq:ucb}
\end{equation}
where the first term is the estimated expected utility and the second term is an uncertainty bonus; $\lambda$ controls the exploration bonus.

This head provides two properties that are critical for deployment.
From the algorithm perspective, it implements an adaptive explore--exploit trade-off: actions with high estimated utility are preferred while actions with high epistemic uncertainty receive a bonus that shrinks as $A_a$ accumulates data.
From our engineering perspective, it is lightweight and predictable in latency (small action set and low-dimensional matrix updates), and it produces first-class observability signals---per-tier counts $|\mathcal{T}_a|$, uncertainty magnitudes $\sqrt{z_t^{\top}A_a^{-1}z_t}$, and tier-histogram drift---which enable rapid diagnosis and conservative rollback in deployment (Section~\ref{sec:payoff}).

\subsubsection{Experience-Oriented Reward Design}
\label{sec:reward_design}

A central challenge in hold control is that our platform cares about overall passenger-driver experience across the funnel rather than a single metric.
For each decision on a \doPair{} at step $t$, the realized outcome may involve multiple stages (\doPair{} broadcast $\rightarrow$ driver acceptance $\rightarrow$ order completion) and heterogeneous failure modes with different user impact.
Moreover, hold influences waiting-related experience through delayed effects.
To train Stage~I with a unified learning signal, we define an \emph{experience-oriented reward} $r_t$ that aggregates outcome utility from heterogeneous impacts and waiting-related penalties while remaining stable for training and deployment.

\paragraph{Outcome utility with heterogeneous funnel impacts.}
Let $y_t \in \mathcal{Y}$ denote the realized outcome category associated with the \doPair{} decision at step $t$ (Section~\ref{sec:pf_outcomes} and Section~\ref{sec:pf_objective}).
We define a base utility that treats successful completion as unit gain and assigns asymmetric penalties to experience-degrading outcomes (Figure~\ref{fig:transition}):
\begin{equation}
r_{\text{base}}(y_t)=
\begin{cases}
+1, & y_t=\CR,\\
-w_b, & y_t=\CBA,\\
-w_p, & y_t=\PCAA,\\
-w_d, & y_t=\DCAA,\\
r_0, & \text{driver non-acceptance},
\end{cases}
\label{eq:base_reward}
\end{equation}
where we set $w_p \approx w_b > w_d \ge 0$, and this hierarchical penalty structure is motivated by the fact that passenger-side negative experiences are usually more difficult to model from our historical data. Therefore, we impose higher penalty weights on passenger-related disruptions to ensure the matching policy remains sensitive to these critical but hard-to-learn outcomes.
In practice, these weights are calibrated from offline statistics (e.g., conditional rates and funnel shares) and business impact considerations.

\paragraph{Tier-dependent cost-effectiveness shaping.}
While Eq.~\eqref{eq:base_reward} provides a consistent utility scale, it does not by itself encourage \emph{selective} holding.
In production, a useful hold policy should improve experience per unit of holding, i.e., it should reduce cancellations and excessive waiting while minimizing unnecessary holding of promising matches. Therefore, we apply a tier- and outcome-dependent shaping term (cost-effectiveness shaping):
\begin{equation}
r_{\text{ce}}(s_t,a_t,y_t)=\eta(a_t,y_t)\cdot r_{\text{base}}(y_t),
\label{eq:ce}
\end{equation}
where $\eta(a_t,y_t)$ is designed to (i) reward higher tiers when they successfully filter experience-degrading outcomes and (ii) penalize higher tiers more strongly when they unnecessarily delay completion.
This shaping makes the learning signal explicitly aligned with ``experience ROI'': it discourages blunt strategies that hold too many candidates and instead favors policies that reserve longer holds for genuinely low-experience cases.

\paragraph{Temporal shaping.}
Hold control is tightly coupled with waiting-related experience.
Let $\text{ETP}_t$ and $\text{PWT}_t$ denote waiting-related signals available at decision time $t$ (Estimated Time of Pickup and Passenger Waiting Time), we construct a mild reward shaping term:
\begin{equation}
r_{\text{time}}(s_t,a_t)=\rho_1 g(\text{ETP}_t)+\rho_2 g(\text{PWT}_t),
\label{eq:time}
\end{equation}
where $g(\cdot)$ is a clipped monotone transform and $(\rho_1,\rho_2)$ are small coefficients.
This term encodes a conservative inductive bias: when waiting is already high, overly aggressive hold is more likely to harm experience, and thus the policy should act more cautiously.
We keep this term deliberately small to avoid destabilizing training or overriding the outcome-based utility.

\paragraph{Final reward.}
The final experience-aware reward is
\begin{equation}
r_t = r_{\text{ce}}(s_t,a_t,y_t) + r_{\text{time}}(s_t,a_t).
\label{eq:final_reward}
\end{equation}
We note that the design of $r_t$ is not to serve as a perfect surrogate for any single target, but to provide a stable learning signal that consistently captures multi-metric experience trade-offs with temporal changes in a way that supports effective and stable deployment.

In practice, jointly updating the encoder and the contextual bandit head can be unstable: encoder updates change the representation space used by the bandit, and exploration during training can amplify simulator imperfections. To obtain a deployment-friendly tiering policy with smooth tier evolution and predictable behavior, we adopt an alternating training scheme with explicit freezes. Parameters for the reward shaping is detailed in Appendix~\ref{sec:appendix}, and the algorithmic details are provided in Appendix~\ref{sec:appendix_alt_training}.

\subsection{Stage II: Guardrail-Constrained Hold-Time Optimization}
\label{sec:stage2}

Stage~II of \method\ converts experience tiers into hold durations by solving a constrained optimization problem based on empirical quantiles.
This stage provides explicit control over unnecessary holding of promising matches and yields monotone, auditable hold schedules that can be versioned and rolled back independently.

\subsubsection{Empirical Quantile Estimation}
\label{sec:quantile}

Given a fixed Stage~I policy, we route \doPair{} samples to tiers and estimate outcome-conditional time distributions.
Let $T$ denote the system-defined hold clock,
for action tier $i$ and outcome category $j\in\mathcal{Y}$, we estimate the corresponding cumulative density function (CDF):
\begin{equation}
\text{CDF}_{i,j}(x)=\mathbb{P}(T\le x \mid a=i, y=j),
\quad
p_{i,j}=\frac{N_{i,j}}{\sum_{i'} N_{i',j}},
\label{eq:cdf_p}
\end{equation}
where $N_{i,j}$ is the number of samples with $(a=i,y=j)$. Quantile summaries are robust and inexpensive to refresh from recent logs.
In production, the estimated quantiles can be easily tracked over time and across cities to detect drift early, and refreshing these summaries enables lightweight, auditable re-calibration of Stage~II schedules without retraining the Stage~I model, supporting safe iteration under non-stationary traffic.

\subsubsection{Constrained Optimization with Unnecessary-Hold Guardrail}
\label{sec:opt}

We choose hold times $\{x_i\}$ to maximize expected utility while bounding the rate of unnecessarily holding completion cases.
We enforce:
(i) a guardrail on completion (unnecessary hold) and
(ii) a monotone schedule for interpretability:
\begin{align}
\max_{\{x_i\}} \quad &
\sum_{i=0}^{K}\sum_{j\in\mathcal{Y}} p_{i,j}\, v_j\, \text{CDF}_{i,j}(x_i) \label{eq:opt_obj}\\
\text{s.t.}\quad &
\sum_{i=1}^{K} p_{i,\CR}\, \text{CDF}_{i,\CR}(x_i)\le \alpha, \label{eq:opt_con}\\
& 0=x_0 \le x_1 \le \cdots \le x_K \le x_{\max}, \label{eq:opt_mono}
\end{align}
where $v_j$ is consistent with the experience utility in Eq.~\eqref{eq:base_reward}, and $\alpha$ is a service guardrail that bounds unnecessary holding of promising matches.
Since $K$ is small in practice, the optimization can be solved efficiently via dynamic programming under the monotonicity constraint, and it produces an explicit lookup table $\{x_i\}$ that is human-reviewable, versioned, and immediately rollback-able.

\subsection{Deployment and Iteration of \method}
\label{sec:e2e}

The deployed \method\ at DiDi is the composition of two decoupled components (shown in Figure~\ref{fig:overview}):
(i) Stage~I tiering policy $\pi_{\theta,\phi}$ that maps each \doPair{} state $s_t$ to an experience tier $a_t\in\mathcal{A}$, and
(ii) Stage~II hold-time table $\{x_i\}_{i=0}^{K}$ obtained by solving the guardrail-constrained optimization in Eq.~\eqref{eq:opt_obj}--\eqref{eq:opt_mono}.
At serving time, the system computes $z_t=f_{\theta}(h_t)$, selects the tier via Eq.~\eqref{eq:ucb}, and executes hold duration $x_{a_t}$.
This design keeps online inference constant-time and makes the execution rule fully explicit and auditable.

One important practical benefit of the two-stage decomposition in \method\ is safe and modular iteration. Stage~I can be updated to improve experience stratification (e.g., better temporal representations or improved bandit calibration), while Stage~II can be refreshed using recent logs to re-estimate empirical quantiles and re-solve the constrained optimization, adjusting conservativeness through the guardrail parameter $\alpha$ without changing Stage~I model weights.
This separation reduces deployment risk, supports rapid rollback, and provides a clean interface between learning and operations.

\section{Experiments}
\label{sec:results}

We present a online A/B evaluation of \method\ on DiDi's large-scale ride-hailing platform.
Our experiments are designed to answer three questions: \textbf{(RQ1)} whether \method\ improves end-to-end trip success and passenger--driver experience under real traffic, \textbf{(RQ2)} which components are essential for the gains, and \textbf{(RQ3)} how the learned policy behaves in online deployment. The application use and payoff of \method\ is detailed in Section~\ref{sec:payoff}.

\subsection{Experimental Setup}
\label{sec:setup}

We evaluate \method\ via an online A/B experiment in DiDi's production ride-hailing matching system.
The control group uses our previous production strategy based on multiple prediction models plus hand-designed threshold rules, which decide whether to hold a candidate \doPair{} and how long to hold.
The treatment group replaces this module with \method: Stage~I assigns each \doPair{} to a discrete experience tier using rich context (Section~\ref{sec:stage1}), and Stage~II maps tiers to a monotone hold-time table computed offline via constrained optimization with explicit service guardrails that limit unnecessary hold of promising matches (Section~\ref{sec:stage2}).

During our experiment, all eligible requests within a 30-minute window are assigned to either control group or treatment group, and the assignment alternates over time. 
We run the experiment from 2025-12-21 to 2026-01-10 (21 days) across five cities in Brazil, covering large-scale production traffic (over $100$k calls per day across the experiment footprint), and aggregate the results over the full experiment window and cities.

We focus on experience-oriented outcomes that hold control is designed to influence.
Our primary metric is the trip completion (\CR) rate and driver income (DI) as the end-to-end measure of trip success.
We also report some important rates at our key funnel stages including PCBA, driver acceptance (DA), PCAA and DCAA, which capture distinct forms of experience degradation and wasted effort.
Finally, because hold directly affects timing, we include waiting-related metrics such as call-acceptance-time (CAT) that records the duration from passenger call to driver acceptance.
\subsection{RQ1: Overall Online Evaluation}
\label{sec:overall}

\begin{table*}[t]
  \centering
  \caption{Online A/B results of \method\ vs.\ our production baseline.
  We report treatment--control deltas (T$-$C). Positive deltas are desirable for metrics marked with $\uparrow$; negative deltas are desirable for metrics marked with $\downarrow$.
  $^*$ indicates statistical significance at $p<0.05$.}
  \label{tab:main_results}
  \begin{tabular}{lccc}
    \toprule
    \textbf{Metric} & \textbf{Delta (T $-$ C)} & \textbf{P-value} & \textbf{Sig.} \\
    \midrule
    \multicolumn{4}{l}{\textbf{Core Measures}} \\
    Trip completion (TC) ratio \hb & \textbf{\posdelta{+0.49\%}} & 0.00273 & $^*$ \\
    Driver income (DI) \hb & \textbf{\posdelta{+0.50\%}} & 0.00014 & $^*$ \\
    \midrule
    \multicolumn{4}{l}{\textbf{Passenger--Driver Experience Measures}} \\
    Driver acceptance (DA) ratio \hb & \textbf{\posdelta{+0.48\%}} & 0.02597 & $^*$ \\
    Passenger cancel before acceptance (\CBA) \lb & \textbf{\posdelta{-1.95\%}} & 0.00402 & $^*$ \\
    Passenger cancel after acceptance (\PCAA) \lb & \textbf{\posdelta{-1.99\%}} & 0.00459 & $^*$ \\
    Driver cancel after acceptance (\DCAA) \lb & \posdelta{-0.92\%} & 0.21310 &  \\
    \midrule
    \multicolumn{4}{l}{\textbf{Ride-Hailing Funnel Measures}} \\
    Call-acceptance time (CAT) \lb & \textbf{\posdelta{-1.80\%}} & 0.04648 & $^*$  \\
    Broadcasted / Called orders \hb & \posdelta{+0.42\%} & 0.05775 &  \\
    Accepted / Broadcasted orders \hb & \textbf{\posdelta{+1.35\%}} & 0.01447 & $^*$ \\
    Accepted / Called orders \hb & \textbf{\posdelta{+1.77\%}} & 0.01143 & $^*$ \\
    \bottomrule
  \end{tabular}
\end{table*}

Table~\ref{tab:main_results} reports the online A/B results of \method\ against our production baseline.
Overall, \method\ delivers consistent improvements on both efficiency and experience.
In particular, we observe statistically significant gains in trip completion and driver income.
The uplift in trip completion suggests that the matches produced under experience-aware hold control lead to a more satisfactory end-to-end trip outcome for both passengers and drivers.
Meanwhile, the improvement in driver income indicates that drivers are more willing to accept the recommended orders after the hold policy filters out low-experience pairings, which directly benefits driver welfare and overall marketplace health.

\paragraph{Experience improvements validate multi-signal reward modeling.}
A key design choice of \method\ is to learn experience tiers with a unified objective that aggregates multiple experience-related signals across the matching funnel (Section~\ref{sec:reward_design}).
Online results support this design: we observe significant reductions in passenger cancellations both before acceptance and after acceptance.
These improvements are particularly important because passenger cancellations are a primary source of experience degradation and wasted matching effort.
Notably, the magnitude and significance of \CBA/\PCAA\ improvements are slightly stronger than \DCAA, which is consistent with our reward weighting that places relatively more emphasis on passenger-side experience.
This suggests a clear path for future online iteration: reward fine-tuning can further strengthen driver-side cancellation control without sacrificing overall experience gains.

\paragraph{More precise holding leads to faster and better matching.}
Beyond outcome metrics, \method\ improves key process indicators in the ride-hailing funnel.
We observe a significant reduction in CAT, indicating that hold is executed more precisely:
by delaying a small subset of low-experience \doPair{} candidates, the system surfaces better opportunities sooner, leading to faster driver response and a more efficient alignment of passenger--driver intent.
Consistent with this interpretation, we also see positive shifts in intermediate conversion metrics: Accepted/Broadcasted and Accepted/Called increases, both statistically significant.
Broadcasted/Called shows a mild increase with borderline significance, suggesting that \method\ does not rely on aggressively suppressing exposure;
instead, improvements primarily come from better conversion after exposure and fewer experience-degrading cancellations.

\begin{table}[t]
  \centering
  \caption{Segment-level online results of \method.
  We report treatment--control deltas (T$-$C) for representative metrics across cities and time segments.}
  \label{tab:segment_results}
  \begin{tabular}{lcccc}
    \toprule
    \textbf{Segment} & \textbf{TC $\uparrow$} & \textbf{DI $\uparrow$} & \textbf{\PCAA{} $\downarrow$} & \textbf{CAT $\downarrow$} \\
    \midrule
    Overall        & +0.49\% & +0.50\% & -1.99\% & -1.80\% \\
    City A         & +0.60\% & +0.63\% & -2.18\% & -1.98\% \\
    City B         & +0.35\% & +0.39\% & -1.84\% & -1.66\% \\
    City C         & +0.42\% & +0.37\% & -1.80\% & -1.63\% \\
    City D         & +0.56\% & +0.46\% & -2.02\% & -1.90\% \\
    City E         & +0.47\% & +0.58\% & -2.06\% & -1.85\% \\
    Peak Hours     & +0.62\% & +0.66\% & -2.24\% & -2.03\% \\
    Off-Peak Hours & +0.31\% & +0.35\% & -1.74\% & -1.56\% \\
    \bottomrule
  \end{tabular}
\end{table}

\paragraph{Robustness across operational segments.}
Aggregate online results may hide heterogeneity across cities and traffic conditions.
To verify that the gains are not driven by a single city or a narrow traffic slice, we further evaluate representative metrics across five cities and peak/off-peak periods; table~\ref{tab:segment_results} reports the segment-level results.
Across all five cities, \method\ consistently improves TC and DI while reducing \PCAA{} and CAT.
The magnitude varies across segments, but the direction remains stable, suggesting that the aggregate gains are not dominated by one city or a specific operational condition.

The time-segment results show a similar pattern.
Both peak and off-peak hours achieve positive gains on TC and DI, together with reductions in \PCAA{} and CAT.
The larger improvements during peak hours are consistent with the intuition that hold decisions become more consequential when marketplace competition and matching pressure are higher.
At the same time, the positive off-peak results indicate that \method\ does not rely solely on peak-hour traffic conditions.
This robustness analysis focuses on online metric consistency, while Section~\ref{sec:pickup_drilldown} further examines the behavioral patterns of hold decisions across pair attributes, temporal conditions, and spatial market regimes.

\begin{table}[t]
  \centering
  \caption{Ablation study of \method\ in online A/B experiments.
  We report treatment--control deltas (T$-$C) for the two core measures.
  Positive deltas are desirable ($\uparrow$).}
  \label{tab:ablation}
  \begin{tabular}{lcc}
    \toprule
    \textbf{Variant} & \textbf{TC Ratio $\uparrow$} & \textbf{DI $\uparrow$} \\
    \midrule
    \method\ (full) & \textbf{\posdelta{+0.49\%}} & \textbf{\posdelta{+0.50\%}} \\
    \midrule
    \multicolumn{3}{l}{\textbf{Stage I ablations}} \\
    w/o cost-effectiveness shaping & \textbf{\negdelta{-0.47\%}} & \textbf{\posdelta{+0.03\%}} \\
    w/o temporal shaping & \textbf{\posdelta{+0.07\%}} & \textbf{\negdelta{-0.29\%}} \\
    w/o any shaping & \textbf{\negdelta{-1.76\%}} & \textbf{\negdelta{-1.85\%}} \\
    \midrule
    \multicolumn{3}{l}{\textbf{Stage II ablations}} \\
    Heuristic hold (only tier $K$, 10s) & \textbf{\negdelta{-1.68\%}} & \textbf{\negdelta{-2.38\%}} \\
    Heuristic hold (only tier $K$, 30s) & \textbf{\negdelta{-2.74\%}} & \textbf{\negdelta{-3.30\%}} \\
    \bottomrule
  \end{tabular}
\end{table}

\subsection{RQ2: Ablation Study}
\label{sec:ablation}

Table~\ref{tab:ablation} reports online ablations of \method\ to isolate the contribution of key design choices in Stage~I (experience-aware reward shaping) and Stage~II (hold-time execution).
All variants are evaluated in the same production A/B setting and we report deltas on the two core measures, TC and DI.

Overall, the full \method\ achieves the best joint improvement on TC and DI, while removing either shaping component in Stage~I degrades performance substantially.
In particular, cost-effectiveness shaping is critical for achieving positive trip completion gains: without it, TC turns negative even though DI becomes nearly neutral, suggesting that the policy loses its ability to trade off experience gains against unnecessary holding in a cost-effective manner. Removing all shaping terms leads to large regressions on both TC and DI, highlighting that the proposed multi-signal reward modeling is essential for the performance of \method.

Stage~II ablations further show that hold-time execution must be guardrail-constrained and tier-specific. Replacing Stage~II with heuristic holds that only act on the worst tier using a fixed duration (10s or 30s) consistently harms both TC and DI.
This indicates that naive ``hold only the worst cases'' is insufficient in production: effective hold control requires calibrated, monotone hold times across experience tiers to avoid over-holding and to maintain balanced improvements for both passengers and drivers. 

\begin{table}[t]
  \centering
  \caption{Sensitivity analysis of Stage~I reward design.}
  \label{tab:reward_sensitivity}
  \begin{tabular}{lcccc}
    \toprule
    \textbf{Reward variant} & \textbf{TC $\uparrow$} & \textbf{DI $\uparrow$} & \textbf{\PCAA{} $\downarrow$} & \textbf{\DCAA{} $\downarrow$} \\
    \midrule
    Symmetric penalty
      & +0.25\%
      & +0.18\%
      & -1.45\%
      & -0.97\% \\
    Driver-heavy penalty
      & +0.33\%
      & +0.43\%
      & -1.20\%
      & -1.06\% \\
    \method
      & +0.49\%
      & +0.50\%
      & -1.99\%
      & -0.92\% \\
    \bottomrule
  \end{tabular}
\end{table}

\paragraph{Reward sensitivity.}
We further test alternative cancellation-penalty structures in Stage~I, including symmetric passenger/driver penalties and driver-side heavier penalties.
The results, reported in Table~\ref{tab:reward_sensitivity}, show that \method\ is not brittle to moderate reward perturbations: all variants remain directionally positive on TC and DI.
However, the penalty structure changes the trade-off across funnel metrics.
Driver-side heavier penalties slightly improve \DCAA{} but weaken passenger-side cancellation reduction and yield smaller TC/DI gains, while the \method\ setting achieves the best joint improvement on TC and DI.

\subsection{RQ3: Behavioral Analysis}
\label{sec:pickup_drilldown}

\begin{figure}[t]
  \centering
  \includegraphics[width=0.48\textwidth]{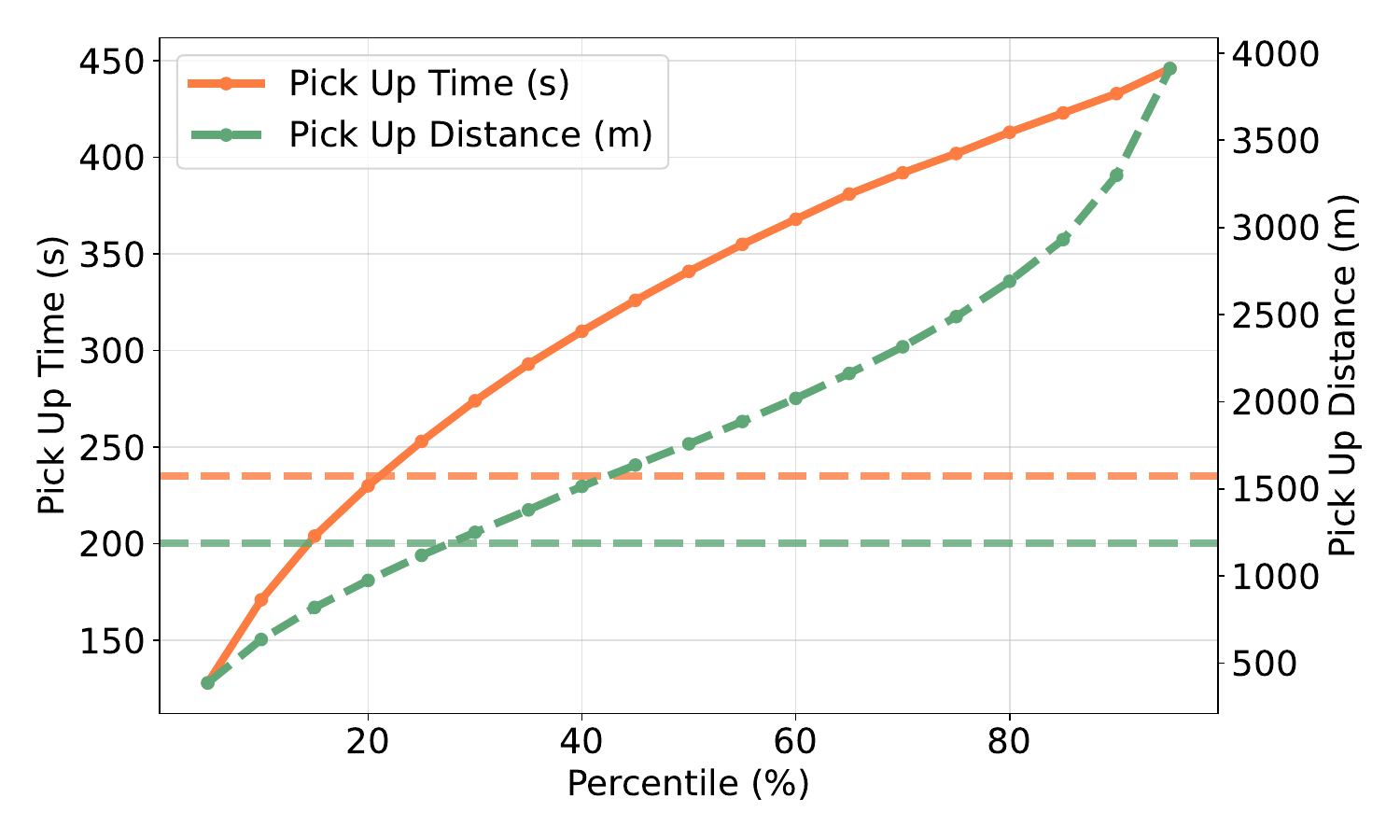}
  \caption{\textbf{Pickup-time/distance profiles of \doPair{}s held by \method; horizontal dashed lines indicate the median pickup time/distance of non-held \doPair{}s.}}
  \label{fig:pickup_quantiles}
\end{figure}

\begin{figure*}[t]
  \centering
  \begin{subfigure}{0.48\textwidth}
    \centering
    \includegraphics[width=\linewidth]{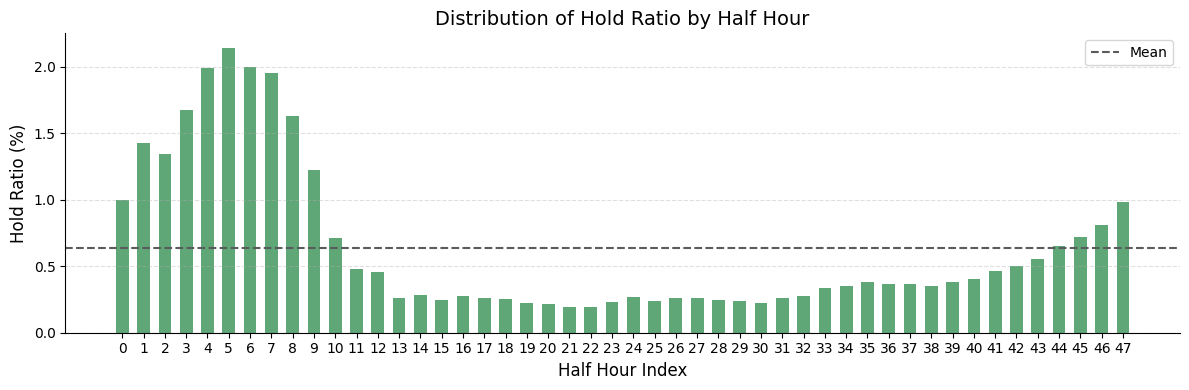}
    \caption{Time-of-day hold ratio.}
    \label{fig:time_bucket}
  \end{subfigure}
  \hfill
  \begin{subfigure}{0.48\textwidth}
    \centering
    \includegraphics[width=\linewidth]{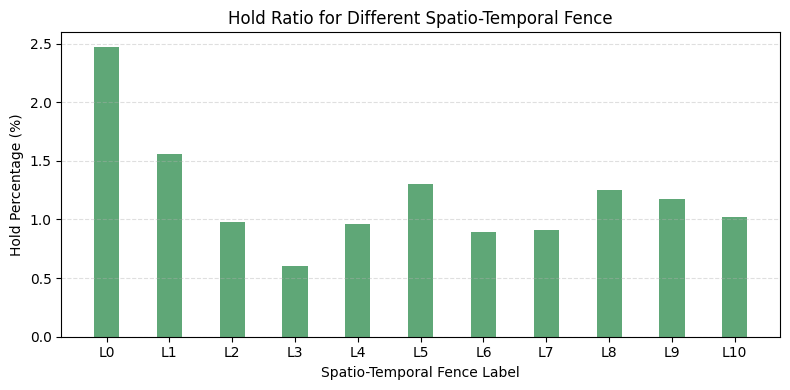}
    \caption{Hold ratio across supply--demand regimes.}
    \label{fig:space_fence}
  \end{subfigure}
  \caption{\textbf{Behavioral analysis of \method\ across temporal and spatial regimes.}
  \method\ holds more frequently in late-night / early-morning periods and in low-density or low-demand-pressure regimes, suggesting that the learned policy adapts to market context rather than applying a uniform hold rate.}
  \label{fig:behavior_regime}
\end{figure*}

Section~\ref{sec:overall} shows that the online gains are directionally consistent across cities and time segments.
We next examine the behavior of the learned hold policy itself.
This analysis complements the segment-level online metrics by asking whether \method\ affects candidates in an interpretable and domain-consistent manner, rather than obtaining gains through indiscriminate exposure suppression or a single hard-coded heuristic.

\paragraph{Pair-level pickup profiles.}
We first examine the pickup time and pickup distance distributions of \doPair{}s held by \method.
Figure~\ref{fig:pickup_quantiles} plots the percentile curves of pickup time and pickup distance for held candidates.
For reference, the horizontal dashed lines indicate the median pickup time and pickup distance of non-held \doPair{}s.

The held candidates tend to have longer pickup time and/or longer pickup distance, as their percentile curves lie above the non-held median at many percentiles.
This pattern is consistent with domain intuition: long pickup time and distance are often associated with worse passenger--driver experience and higher cancellation propensity.
Delaying such candidates can therefore create opportunities for more mutually satisfactory matches.

Importantly, the behavior is not reducible to a single pickup-time or pickup-distance threshold.
Some held \doPair{}s still have relatively short pickup time or distance, while not all long-pickup candidates are necessarily held.
This is consistent with the design of \method: Stage~I assigns an experience tier by aggregating multiple funnel and waiting-related signals, and Stage~II executes tier-specific hold durations obtained from constrained optimization.
Therefore, the learned policy performs calibrated experience-aware filtering rather than applying a one-size-fits-all rule.

\paragraph{Temporal behavior.}
We next examine whether the hold policy behaves sensibly across diurnal cycles.
Figure~\ref{fig:behavior_regime}(a) reports the ratio of held candidates over half-hour time buckets.
The hold ratio is higher during late-night and early-morning periods, while remaining lower and relatively stable during daytime.
This pattern is consistent with operational intuition that off-peak periods often contain more uncertain or lower-connectivity matching opportunities.
The result suggests that \method\ adapts its hold intensity to temporal market conditions, instead of applying a uniform hold rate throughout the day.

\paragraph{Spatial regime behavior.}
Finally, we analyze how hold decisions vary across local supply--demand regimes.
We stratify candidates by spatiotemporal fence level, which reflects demand pressure, and by order density, which reflects local connectivity and available matching opportunities.
As shown in Figure~\ref{fig:behavior_regime}(b), \method\ holds more frequently in low-demand-pressure and low-density regimes (L0-L4), where risky \doPair{}s are generally more common.
In medium-to-high demand regimes (L5-L10), the hold ratio becomes more moderate and varies with local density, indicating that the policy does not simply map regime labels to fixed hold actions.
Within comparable demand-pressure groups, the hold ratio generally decreases as density increases (e.g., L5-L7, L8-L10), suggesting that the policy becomes less conservative with higher local connectivity.

Together, these behavioral analyses show that \method\ makes hold decisions that are interpretable at multiple levels.
At the pair level, it tends to delay candidates with higher pickup burden.
At the temporal and spatial levels, it increases hold intensity in regimes where matching uncertainty is higher, while avoiding uniformly aggressive holding in better-connected markets.
These patterns support our interpretation that the online gains are driven by selective risk reduction and calibrated hold execution.

\subsection{Application Use and Payoff}
\label{sec:payoff}

\method\ has been deployed in DiDi's production ride-hailing matching system and is currently running to serve large-scale real traffic in Brazil, becoming an active component of DiDi's marketplace policy stack.
In production operation, \method\ delivers sustained gains on trip completion and passenger--driver experience, demonstrating that experience-aware real-time hold control can create measurable end-to-end benefits.

\paragraph{From offline design to production deployment.}
Offline evaluation played a central role in turning \method\ into a shippable policy.
For Stage~I, we iterated the reward design and used a \emph{cost-effectiveness score} as the primary model-selection signal to guide reward shaping and weight tuning~\ref{sec:appendix}.
This score improved from $0.89$ to $1.63$, and further to $2.62$ after introducing cost-effectiveness and temporal shaping, indicating that the learned policy increasingly reduces experience-degrading outcomes per unit of unnecessary holding.
For Stage~II, we explored multiple constraint configurations and found that setting the guardrail parameter to $\alpha=0.01$ yields the best balance in our offline simulation and also translates most reliably to online A/B performance, aligning the hold-time schedule with practical service constraints while preserving experience gains.

\paragraph{Monitoring, rollback, and serving efficiency.}

To ensure stable production performance during experimentation and deployment, we built end-to-end monitoring around several critical checkpoints. We continuously track the distribution of Stage~I outputs (the fraction of \doPair{}s assigned to each experience tier) and alert on abrupt shifts that may indicate feature issues. We also monitor Stage~II optimization outputs (e.g., the monotone hold-time table and constraint satisfaction) and key online service indicators, with automatic alarm and rollback mechanisms to protect against regressions. From an efficiency perspective, engineering profiling shows that \method\ has a computational footprint comparable to the existing predictive-model-based hold module: Stage~I adds a bounded-cost embedding inference and discrete selection, while Stage~II is an offline table lookup at runtime. Stress tests under peak traffic further confirm that the method meets the latency and throughput requirements of real-time, large-scale inference in production.
\section{Related Work}
\label{sec:related_work}

\paragraph{Ride-hailing dispatch and matching systems.}
Large-scale ride-hailing platforms require real-time, high-throughput decision making over a dynamic bipartite market under tight latency and reliability constraints.
Prior work has studied learning-augmented dispatch and matching mechanisms at industrial scale, including approaches that combine learning with downstream planning or optimization to improve global efficiency and user experience \cite{xu2018large, liu2022deep}.
A complementary line of work focuses on modeling pairwise matching quality, e.g., predicting the success probability of passenger--driver pairs under multi-view features and deployment constraints \cite{wang2021secure, teusch2023systematic, suhr2019two}.
More broadly, research in operations and marketplaces has analyzed the coupling between matching, waiting time, and platform objectives such as experience and efficiency \cite{yan2020dynamic, taylor2024shared, ke2021equilibrium}, which motivates decision modules that explicitly manage waiting-related outcomes and cancellations.

\paragraph{Experience quality and funnel-aware objectives.}
Cancellations in ride-hailing are heterogeneous and occur at different stages of the funnel, reflecting distinct sources of experience degradation and operational waste~\cite{gangadharaiah2025understanding, xu2023modeling, feng2021we}.
While many industrial systems rely on predictive models (e.g., completion/cancellation estimators) to drive heuristic thresholding policies, such designs can struggle with multi-objective coordination and cascading errors when multiple models are chained~\cite{supian2024ride, guomulti, supian2025mathematical}.
Recent platform studies highlight the importance of accounting for two-sided effects and system-level externalities in repeated matching markets, including fairness and long-term outcomes beyond immediate conversion \cite{suhr2019two}.
Contextual bandits and constrained optimization have also been explored for dispatch and other marketplace controls~\cite{qin2021optimizing, yuan2022reinforcement, liu2022deep, liu2025adaptive}, but production deployment often faces challenges including delayed effects, distribution shift, and safety constraints.

\section{Conclusion}
\label{sec:conclusion}

This paper presented \method, a deployable two-stage framework for {experience-aware real-time hold control} in large-scale ride-hailing matching at DiDi global markets.
The key idea of \method\ is to decouple experience-aware pair assessment from hold-time execution under explicit service guardrails:
Stage~I learns discrete and interpretable experience tiers from sequential context using a robust decision model, and Stage~II converts tiers into a monotone hold-time schedule via constrained optimization over empirical quantiles. Through online A/B experiments and production deployment, we show that \method\ improves trip success and driver welfare while reducing experience-degrading cancellations and improving funnel efficiency, providing real-world evidence that properly calibrated hold control can benefit both sides of the marketplace at DiDi.

\newpage


\begin{thebibliography}{33}

  
  \ifx \showCODEN    \undefined \def \showCODEN     #1{\unskip}     \fi
  \ifx \showISBNx    \undefined \def \showISBNx     #1{\unskip}     \fi
  \ifx \showISBNxiii \undefined \def \showISBNxiii  #1{\unskip}     \fi
  \ifx \showISSN     \undefined \def \showISSN      #1{\unskip}     \fi
  \ifx \showLCCN     \undefined \def \showLCCN      #1{\unskip}     \fi
  \ifx \shownote     \undefined \def \shownote      #1{#1}          \fi
  \ifx \showarticletitle \undefined \def \showarticletitle #1{#1}   \fi
  \ifx \showURL      \undefined \def \showURL       {\relax}        \fi
  \providecommand\bibfield[2]{#2}
  \providecommand\bibinfo[2]{#2}
  \providecommand\natexlab[1]{#1}
  \providecommand\showeprint[2][]{arXiv:#2}
  
  \bibitem[Af{\`e}che et~al\mbox{.}(2023)]%
          {afeche2023ride}
  \bibfield{author}{\bibinfo{person}{Philipp Af{\`e}che}, \bibinfo{person}{Zhe Liu}, {and} \bibinfo{person}{Costis Maglaras}.} \bibinfo{year}{2023}\natexlab{}.
  \newblock \showarticletitle{Ride-hailing networks with strategic drivers: The impact of platform control capabilities on performance}.
  \newblock \bibinfo{journal}{\emph{Manufacturing \& Service Operations Management}} \bibinfo{volume}{25}, \bibinfo{number}{5} (\bibinfo{year}{2023}), \bibinfo{pages}{1890--1908}.
  \newblock
  
  
  \bibitem[Barbour and Luiz(2019)]%
          {barbour2019embracing}
  \bibfield{author}{\bibinfo{person}{Olivia Barbour} {and} \bibinfo{person}{John Luiz}.} \bibinfo{year}{2019}\natexlab{}.
  \newblock \showarticletitle{Embracing solutions-driven innovation to address institutional voids: The case of uber and the middle of the pyramid}.
  \newblock \bibinfo{journal}{\emph{California Management Review}} \bibinfo{volume}{62}, \bibinfo{number}{1} (\bibinfo{year}{2019}), \bibinfo{pages}{31--52}.
  \newblock
  
  
  \bibitem[Chen et~al\mbox{.}(2021)]%
          {chen2021short}
  \bibfield{author}{\bibinfo{person}{Long Chen}, \bibinfo{person}{Piyushimita Thakuriah}, {and} \bibinfo{person}{Konstantinos Ampountolas}.} \bibinfo{year}{2021}\natexlab{}.
  \newblock \showarticletitle{Short-term prediction of demand for ride-hailing services: A deep learning approach}.
  \newblock \bibinfo{journal}{\emph{Journal of Big Data Analytics in Transportation}} \bibinfo{volume}{3}, \bibinfo{number}{2} (\bibinfo{year}{2021}), \bibinfo{pages}{175--195}.
  \newblock
  
  
  \bibitem[Chu et~al\mbox{.}(2011)]%
          {chu2011contextual}
  \bibfield{author}{\bibinfo{person}{Wei Chu}, \bibinfo{person}{Lihong Li}, \bibinfo{person}{Lev Reyzin}, {and} \bibinfo{person}{Robert Schapire}.} \bibinfo{year}{2011}\natexlab{}.
  \newblock \showarticletitle{Contextual bandits with linear payoff functions}. In \bibinfo{booktitle}{\emph{Proceedings of the fourteenth international conference on artificial intelligence and statistics}}. JMLR Workshop and Conference Proceedings, \bibinfo{pages}{208--214}.
  \newblock
  
  
  \bibitem[Feng et~al\mbox{.}(2021)]%
          {feng2021we}
  \bibfield{author}{\bibinfo{person}{Guiyun Feng}, \bibinfo{person}{Guangwen Kong}, {and} \bibinfo{person}{Zizhuo Wang}.} \bibinfo{year}{2021}\natexlab{}.
  \newblock \showarticletitle{We are on the way: Analysis of on-demand ride-hailing systems}.
  \newblock \bibinfo{journal}{\emph{Manufacturing \& Service Operations Management}} \bibinfo{volume}{23}, \bibinfo{number}{5} (\bibinfo{year}{2021}), \bibinfo{pages}{1237--1256}.
  \newblock
  
  
  \bibitem[Gangadharaiah(2025)]%
          {gangadharaiah2025understanding}
  \bibfield{author}{\bibinfo{person}{Rakesh Gangadharaiah}.} \bibinfo{year}{2025}\natexlab{}.
  \newblock \showarticletitle{Understanding and Modeling Pooled Rideshare Acceptance: Influential Factors, Preferred User Experiences, and Implications}.
  \newblock  (\bibinfo{year}{2025}).
  \newblock
  
  
  \bibitem[Guo et~al\mbox{.}({[n.\,d.]})]%
          {guomulti}
  \bibfield{author}{\bibinfo{person}{Yuhan Guo}, \bibinfo{person}{Wenhua Li}, \bibinfo{person}{Linfan Xiao}, \bibinfo{person}{Hamid Allaoui}, {et~al\mbox{.}}} \bibinfo{year}{[n.\,d.]}\natexlab{}.
  \newblock \showarticletitle{Multi-Objective Dispatching Strategy of Autonomous Service Vehicles in Ride-Hailing Based on an Interpretable Prediction Model}.
  \newblock \bibinfo{journal}{\emph{Hamid, Multi-Objective Dispatching Strategy of Autonomous Service Vehicles in Ride-Hailing Based on an Interpretable Prediction Model}} (\bibinfo{year}{[n.\,d.]}).
  \newblock
  
  
  \bibitem[Ke et~al\mbox{.}(2021)]%
          {ke2021equilibrium}
  \bibfield{author}{\bibinfo{person}{Jintao Ke}, \bibinfo{person}{Zheng Zhu}, \bibinfo{person}{Hai Yang}, {and} \bibinfo{person}{Qiaochu He}.} \bibinfo{year}{2021}\natexlab{}.
  \newblock \showarticletitle{Equilibrium analyses and operational designs of a coupled market with substitutive and complementary ride-sourcing services to public transits}.
  \newblock \bibinfo{journal}{\emph{Transportation Research Part E: Logistics and Transportation Review}}  \bibinfo{volume}{148} (\bibinfo{year}{2021}), \bibinfo{pages}{102236}.
  \newblock
  
  
  \bibitem[Kotary et~al\mbox{.}(2021)]%
          {kotary2021end}
  \bibfield{author}{\bibinfo{person}{James Kotary}, \bibinfo{person}{Ferdinando Fioretto}, \bibinfo{person}{Pascal Van~Hentenryck}, {and} \bibinfo{person}{Bryan Wilder}.} \bibinfo{year}{2021}\natexlab{}.
  \newblock \showarticletitle{End-to-end constrained optimization learning: A survey}.
  \newblock \bibinfo{journal}{\emph{arXiv preprint arXiv:2103.16378}} (\bibinfo{year}{2021}).
  \newblock
  
  
  \bibitem[Lee et~al\mbox{.}(2022)]%
          {lee2022impact}
  \bibfield{author}{\bibinfo{person}{Kyunghee Lee}, \bibinfo{person}{Qianran Jin}, \bibinfo{person}{Animesh Animesh}, {and} \bibinfo{person}{Jui Ramaprasad}.} \bibinfo{year}{2022}\natexlab{}.
  \newblock \showarticletitle{Impact of ride-hailing services on transportation mode choices: evidence from traffic and transit ridership}.
  \newblock \bibinfo{journal}{\emph{MIS Quarterly}} \bibinfo{volume}{46}, \bibinfo{number}{4} (\bibinfo{year}{2022}), \bibinfo{pages}{1875--1900}.
  \newblock
  
  
  \bibitem[Liang et~al\mbox{.}(2022)]%
          {liang2022survey}
  \bibfield{author}{\bibinfo{person}{Jing Liang}, \bibinfo{person}{Xuanxuan Ban}, \bibinfo{person}{Kunjie Yu}, \bibinfo{person}{Boyang Qu}, \bibinfo{person}{Kangjia Qiao}, \bibinfo{person}{Caitong Yue}, \bibinfo{person}{Ke Chen}, {and} \bibinfo{person}{Kay~Chen Tan}.} \bibinfo{year}{2022}\natexlab{}.
  \newblock \showarticletitle{A survey on evolutionary constrained multiobjective optimization}.
  \newblock \bibinfo{journal}{\emph{IEEE Transactions on Evolutionary Computation}} \bibinfo{volume}{27}, \bibinfo{number}{2} (\bibinfo{year}{2022}), \bibinfo{pages}{201--221}.
  \newblock
  
  
  \bibitem[Liu et~al\mbox{.}(2025)]%
          {liu2025adaptive}
  \bibfield{author}{\bibinfo{person}{Xu Liu}, \bibinfo{person}{Yiqiang Lu}, \bibinfo{person}{Jian Liu}, \bibinfo{person}{Tianyi Zhang}, \bibinfo{person}{Weiqiang Wang}, \bibinfo{person}{Qian Liu}, {and} \bibinfo{person}{Shuai Li}.} \bibinfo{year}{2025}\natexlab{}.
  \newblock \showarticletitle{Adaptive Merchant-Centric Risk Control via Unbiased Decision-Making and Dynamic Optimization in E-Commerce}. In \bibinfo{booktitle}{\emph{Proceedings of the AAAI Conference on Artificial Intelligence}}, Vol.~\bibinfo{volume}{39}. \bibinfo{pages}{28783--28791}.
  \newblock
  
  
  \bibitem[Liu et~al\mbox{.}(2022)]%
          {liu2022deep}
  \bibfield{author}{\bibinfo{person}{Yang Liu}, \bibinfo{person}{Fanyou Wu}, \bibinfo{person}{Cheng Lyu}, \bibinfo{person}{Shen Li}, \bibinfo{person}{Jieping Ye}, {and} \bibinfo{person}{Xiaobo Qu}.} \bibinfo{year}{2022}\natexlab{}.
  \newblock \showarticletitle{Deep dispatching: A deep reinforcement learning approach for vehicle dispatching on online ride-hailing platform}.
  \newblock \bibinfo{journal}{\emph{Transportation Research Part E: Logistics and Transportation Review}}  \bibinfo{volume}{161} (\bibinfo{year}{2022}), \bibinfo{pages}{102694}.
  \newblock
  
  
  \bibitem[Naumov and Keith(2023)]%
          {naumov2023optimizing}
  \bibfield{author}{\bibinfo{person}{Sergey Naumov} {and} \bibinfo{person}{David Keith}.} \bibinfo{year}{2023}\natexlab{}.
  \newblock \showarticletitle{Optimizing the economic and environmental benefits of ride-hailing and pooling}.
  \newblock \bibinfo{journal}{\emph{Production and Operations Management}} \bibinfo{volume}{32}, \bibinfo{number}{3} (\bibinfo{year}{2023}), \bibinfo{pages}{904--929}.
  \newblock
  
  
  \bibitem[Qin et~al\mbox{.}(2021)]%
          {qin2021optimizing}
  \bibfield{author}{\bibinfo{person}{Guoyang Qin}, \bibinfo{person}{Qi Luo}, \bibinfo{person}{Yafeng Yin}, \bibinfo{person}{Jian Sun}, {and} \bibinfo{person}{Jieping Ye}.} \bibinfo{year}{2021}\natexlab{}.
  \newblock \showarticletitle{Optimizing matching time intervals for ride-hailing services using reinforcement learning}.
  \newblock \bibinfo{journal}{\emph{Transportation Research Part C: Emerging Technologies}}  \bibinfo{volume}{129} (\bibinfo{year}{2021}), \bibinfo{pages}{103239}.
  \newblock
  
  
  \bibitem[Somalwar et~al\mbox{.}(2025)]%
          {somalwar2025learning}
  \bibfield{author}{\bibinfo{person}{Anne Somalwar}, \bibinfo{person}{Bruce~D Lee}, \bibinfo{person}{George~J Pappas}, {and} \bibinfo{person}{Nikolai Matni}.} \bibinfo{year}{2025}\natexlab{}.
  \newblock \showarticletitle{Learning with Imperfect Models: When Multi-step Prediction Mitigates Compounding Error}.
  \newblock \bibinfo{journal}{\emph{arXiv preprint arXiv:2504.01766}} (\bibinfo{year}{2025}).
  \newblock
  
  
  \bibitem[S{\"u}hr et~al\mbox{.}(2019)]%
          {suhr2019two}
  \bibfield{author}{\bibinfo{person}{Tom S{\"u}hr}, \bibinfo{person}{Asia~J Biega}, \bibinfo{person}{Meike Zehlike}, \bibinfo{person}{Krishna~P Gummadi}, {and} \bibinfo{person}{Abhijnan Chakraborty}.} \bibinfo{year}{2019}\natexlab{}.
  \newblock \showarticletitle{Two-sided fairness for repeated matchings in two-sided markets: A case study of a ride-hailing platform}. In \bibinfo{booktitle}{\emph{Proceedings of the 25th ACM SIGKDD international conference on knowledge discovery \& data mining}}. \bibinfo{pages}{3082--3092}.
  \newblock
  
  
  \bibitem[Supian et~al\mbox{.}(2024)]%
          {supian2024ride}
  \bibfield{author}{\bibinfo{person}{Sudradjat Supian}, \bibinfo{person}{Subiyanto}, \bibinfo{person}{Tubagus~Robbi Megantara}, {and} \bibinfo{person}{Abdul~Talib Bon}.} \bibinfo{year}{2024}\natexlab{}.
  \newblock \showarticletitle{Ride-Hailing Matching with Uncertain Travel Time: A Novel Interval-Valued Fuzzy Multi-Objective Linear Programming Approach}.
  \newblock \bibinfo{journal}{\emph{Mathematics}} \bibinfo{volume}{12}, \bibinfo{number}{9} (\bibinfo{year}{2024}), \bibinfo{pages}{1355}.
  \newblock
  
  
  \bibitem[Supian et~al\mbox{.}(2025)]%
          {supian2025mathematical}
  \bibfield{author}{\bibinfo{person}{Sudradjat Supian}, \bibinfo{person}{Subiyanto Subiyanto}, \bibinfo{person}{Sisilia Sylviani}, \bibinfo{person}{Tubagus~Robbi Megantara}, \bibinfo{person}{Abdul~Talib Bon}, {and} \bibinfo{person}{Vasile Preda}.} \bibinfo{year}{2025}\natexlab{}.
  \newblock \showarticletitle{Mathematical Modeling of Ride-Hailing Matching Considering Uncertain User and Driver Preferences: Interval-Valued Fuzzy Approach}.
  \newblock \bibinfo{journal}{\emph{Mathematics}} \bibinfo{volume}{13}, \bibinfo{number}{3} (\bibinfo{year}{2025}), \bibinfo{pages}{371}.
  \newblock
  
  
  \bibitem[Taylor(2024)]%
          {taylor2024shared}
  \bibfield{author}{\bibinfo{person}{Terry~A Taylor}.} \bibinfo{year}{2024}\natexlab{}.
  \newblock \showarticletitle{Shared-ride efficiency of ride-hailing platforms}.
  \newblock \bibinfo{journal}{\emph{Manufacturing \& Service Operations Management}} \bibinfo{volume}{26}, \bibinfo{number}{5} (\bibinfo{year}{2024}), \bibinfo{pages}{1945--1961}.
  \newblock
  
  
  \bibitem[Teusch et~al\mbox{.}(2023)]%
          {teusch2023systematic}
  \bibfield{author}{\bibinfo{person}{Julian Teusch}, \bibinfo{person}{Jan~Niklas Gremmel}, \bibinfo{person}{Christian Koetsier}, \bibinfo{person}{Fatema~Tuj Johora}, \bibinfo{person}{Monika Sester}, \bibinfo{person}{David~M Woisetschl{\"a}ger}, {and} \bibinfo{person}{J{\"o}rg~P M{\"u}ller}.} \bibinfo{year}{2023}\natexlab{}.
  \newblock \showarticletitle{A systematic literature review on machine learning in shared mobility}.
  \newblock \bibinfo{journal}{\emph{IEEE Open Journal of Intelligent Transportation Systems}}  \bibinfo{volume}{4} (\bibinfo{year}{2023}), \bibinfo{pages}{870--899}.
  \newblock
  
  
  \bibitem[Tirachini(2020)]%
          {tirachini2020ride}
  \bibfield{author}{\bibinfo{person}{Alejandro Tirachini}.} \bibinfo{year}{2020}\natexlab{}.
  \newblock \showarticletitle{Ride-hailing, travel behaviour and sustainable mobility: an international review}.
  \newblock \bibinfo{journal}{\emph{Transportation}} \bibinfo{volume}{47}, \bibinfo{number}{4} (\bibinfo{year}{2020}), \bibinfo{pages}{2011--2047}.
  \newblock
  
  
  \bibitem[Tu(2024)]%
          {tu2024towards}
  \bibfield{author}{\bibinfo{person}{Yuanjie Tu}.} \bibinfo{year}{2024}\natexlab{}.
  \newblock \emph{\bibinfo{title}{Towards sustainability: decoding ridehailing drivers' and passengers' behaviors}}.
  \newblock \bibinfo{thesistype}{Ph.\,D. Dissertation}. \bibinfo{school}{University of Washington}.
  \newblock
  
  
  \bibitem[Vaswani et~al\mbox{.}(2017)]%
          {vaswani2017attention}
  \bibfield{author}{\bibinfo{person}{Ashish Vaswani}, \bibinfo{person}{Noam Shazeer}, \bibinfo{person}{Niki Parmar}, \bibinfo{person}{Jakob Uszkoreit}, \bibinfo{person}{Llion Jones}, \bibinfo{person}{Aidan~N Gomez}, \bibinfo{person}{{\L}ukasz Kaiser}, {and} \bibinfo{person}{Illia Polosukhin}.} \bibinfo{year}{2017}\natexlab{}.
  \newblock \showarticletitle{Attention is all you need}.
  \newblock \bibinfo{journal}{\emph{Advances in neural information processing systems}}  \bibinfo{volume}{30} (\bibinfo{year}{2017}).
  \newblock
  
  
  \bibitem[Wang et~al\mbox{.}(2021)]%
          {wang2021secure}
  \bibfield{author}{\bibinfo{person}{Yuandong Wang}, \bibinfo{person}{Hongzhi Yin}, \bibinfo{person}{Lian Wu}, \bibinfo{person}{Tong Chen}, {and} \bibinfo{person}{Chunyang Liu}.} \bibinfo{year}{2021}\natexlab{}.
  \newblock \showarticletitle{Secure your ride: Real-time matching success rate prediction for passenger-driver pairs}.
  \newblock \bibinfo{journal}{\emph{IEEE Transactions on Knowledge and Data Engineering}} \bibinfo{volume}{35}, \bibinfo{number}{3} (\bibinfo{year}{2021}), \bibinfo{pages}{3059--3071}.
  \newblock
  
  
  \bibitem[Wen et~al\mbox{.}(2024)]%
          {wen2024survey}
  \bibfield{author}{\bibinfo{person}{Dacheng Wen}, \bibinfo{person}{Yupeng Li}, {and} \bibinfo{person}{Francis~CM Lau}.} \bibinfo{year}{2024}\natexlab{}.
  \newblock \showarticletitle{A survey of machine learning-based ride-hailing planning}.
  \newblock \bibinfo{journal}{\emph{IEEE Transactions on Intelligent Transportation Systems}} \bibinfo{volume}{25}, \bibinfo{number}{6} (\bibinfo{year}{2024}), \bibinfo{pages}{4734--4753}.
  \newblock
  
  
  \bibitem[Xu(2023)]%
          {xu2023modeling}
  \bibfield{author}{\bibinfo{person}{Kai Xu}.} \bibinfo{year}{2023}\natexlab{}.
  \newblock \emph{\bibinfo{title}{Modeling and Management of Ridesourcing Services with Order Cancellation and Platform Collaboration}}.
  \newblock \bibinfo{thesistype}{Ph.\,D. Dissertation}. \bibinfo{school}{UNSW Sydney}.
  \newblock
  
  
  \bibitem[Xu et~al\mbox{.}(2018)]%
          {xu2018large}
  \bibfield{author}{\bibinfo{person}{Zhe Xu}, \bibinfo{person}{Zhixin Li}, \bibinfo{person}{Qingwen Guan}, \bibinfo{person}{Dingshui Zhang}, \bibinfo{person}{Qiang Li}, \bibinfo{person}{Junxiao Nan}, \bibinfo{person}{Chunyang Liu}, \bibinfo{person}{Wei Bian}, {and} \bibinfo{person}{Jieping Ye}.} \bibinfo{year}{2018}\natexlab{}.
  \newblock \showarticletitle{Large-scale order dispatch in on-demand ride-hailing platforms: A learning and planning approach}. In \bibinfo{booktitle}{\emph{Proceedings of the 24th ACM SIGKDD international conference on knowledge discovery \& data mining}}. \bibinfo{pages}{905--913}.
  \newblock
  
  
  \bibitem[Yan et~al\mbox{.}(2020)]%
          {yan2020dynamic}
  \bibfield{author}{\bibinfo{person}{Chiwei Yan}, \bibinfo{person}{Helin Zhu}, \bibinfo{person}{Nikita Korolko}, {and} \bibinfo{person}{Dawn Woodard}.} \bibinfo{year}{2020}\natexlab{}.
  \newblock \showarticletitle{Dynamic pricing and matching in ride-hailing platforms}.
  \newblock \bibinfo{journal}{\emph{Naval Research Logistics (NRL)}} \bibinfo{volume}{67}, \bibinfo{number}{8} (\bibinfo{year}{2020}), \bibinfo{pages}{705--724}.
  \newblock
  
  
  \bibitem[Yang et~al\mbox{.}(2020)]%
          {yang2020phase}
  \bibfield{author}{\bibinfo{person}{Bo Yang}, \bibinfo{person}{Shen Ren}, \bibinfo{person}{Erika~Fille Legara}, \bibinfo{person}{Zengxiang Li}, \bibinfo{person}{Edward~YX Ong}, \bibinfo{person}{Louis Lin}, {and} \bibinfo{person}{Christopher Monterola}.} \bibinfo{year}{2020}\natexlab{}.
  \newblock \showarticletitle{Phase transition in taxi dynamics and impact of ridesharing}.
  \newblock \bibinfo{journal}{\emph{Transportation Science}} \bibinfo{volume}{54}, \bibinfo{number}{1} (\bibinfo{year}{2020}), \bibinfo{pages}{250--273}.
  \newblock
  
  
  \bibitem[Yatnalkar(2019)]%
          {yatnalkar2019machine}
  \bibfield{author}{\bibinfo{person}{Govind~Pramod Yatnalkar}.} \bibinfo{year}{2019}\natexlab{}.
  \newblock \showarticletitle{A Machine Learning Recommender Model for Ride Sharing Based on Rider Characteristics and User Threshold Time}.
  \newblock  (\bibinfo{year}{2019}).
  \newblock
  
  
  \bibitem[Yuan et~al\mbox{.}(2022)]%
          {yuan2022reinforcement}
  \bibfield{author}{\bibinfo{person}{Enpeng Yuan}, \bibinfo{person}{Wenbo Chen}, {and} \bibinfo{person}{Pascal Van~Hentenryck}.} \bibinfo{year}{2022}\natexlab{}.
  \newblock \showarticletitle{Reinforcement learning from optimization proxy for ride-hailing vehicle relocation}.
  \newblock \bibinfo{journal}{\emph{Journal of Artificial Intelligence Research}}  \bibinfo{volume}{75} (\bibinfo{year}{2022}), \bibinfo{pages}{985--1002}.
  \newblock
  
  
  \bibitem[Zhang et~al\mbox{.}(2017)]%
          {zhang2017taxi}
  \bibfield{author}{\bibinfo{person}{Lingyu Zhang}, \bibinfo{person}{Tao Hu}, \bibinfo{person}{Yue Min}, \bibinfo{person}{Guobin Wu}, \bibinfo{person}{Junying Zhang}, \bibinfo{person}{Pengcheng Feng}, \bibinfo{person}{Pinghua Gong}, {and} \bibinfo{person}{Jieping Ye}.} \bibinfo{year}{2017}\natexlab{}.
  \newblock \showarticletitle{A taxi order dispatch model based on combinatorial optimization}. In \bibinfo{booktitle}{\emph{Proceedings of the 23rd ACM SIGKDD international conference on knowledge discovery and data mining}}. \bibinfo{pages}{2151--2159}.
  \newblock
  
  
  \end{thebibliography}


\appendix









\appendix

\section{Additional Details}
\label{sec:appendix}

\subsection{Reward Table and Shaping Details}
\label{sec:appendix_reward}

This appendix provides a concrete instantiation of the experience-oriented reward used in Stage~I (Section~\ref{sec:reward_design}).
The reward is composed of (i) a base outcome utility, (ii) tier-dependent cost-effectiveness shaping, and (iii) mild temporal shaping.
Throughout, the action $a\in\{0,1,2,3\}$ denotes the predicted \emph{experience tier}, where larger $a$ corresponds to lower expected experience and thus a more conservative holding intent.

\subsubsection{Base reward table}
\label{sec:appendix_base_reward}

We define a categorical outcome $y$ for each \doPair{} that reflects the realized end state in the funnel:
completion (\CR), passenger cancel before acceptance (\CBA), passenger cancel after acceptance (\PCAA), driver cancel after acceptance (\DCAA), and \emph{other} (e.g., non-response or undefined terminals).
The base utility $r_{\text{base}}(y)$ sets completion as the unit benefit and assigns asymmetric penalties to experience-degrading outcomes.
A reasonable instantiation (used in our offline iteration) is shown in Table~\ref{tab:reward_base}.

\begin{table}[h]
  \centering
  \caption{Base reward table $r_{\text{base}}(y)$.}
  \label{tab:reward_base}
  \begin{tabular}{lc}
    \toprule
    Outcome $y$ & $r_{\text{base}}(y)$ \\
    \midrule
    Completion (\CR) & $+1.00$ \\
    Passenger cancel before acceptance (\CBA) & $-0.60$ \\
    Passenger cancel after acceptance (\PCAA) & $-1.20$ \\
    Driver cancel after acceptance (\DCAA) & $-0.80$ \\
    Other / non-response & $-0.20$ \\
    \bottomrule
  \end{tabular}
\end{table}

The table reflects two principles: (i) passenger cancellations after acceptance are typically the most harmful (wasted pickup effort plus poor passenger experience), and (ii) non-response / other outcomes are mildly negative to discourage inefficient pairings.

\subsubsection{Tier-dependent cost-effectiveness shaping}
\label{sec:appendix_ce_shaping}

The base reward alone does not encode \emph{how aggressively} the policy should hold; in production, we want the policy to reduce experience-degrading outcomes while limiting unnecessary holding of promising matches.
We therefore apply a multiplicative shaping term:
\begin{equation}
r_{\text{ce}}(s,a,y) \;=\; \eta(a,y)\cdot r_{\text{base}}(y),
\label{eq:appendix_ce}
\end{equation}
where $\eta(a,y)$ adjusts the utility to reflect effectiveness across tiers.

\paragraph{Design intuition.}
For higher tiers, it is desirable to (i) assign larger positive credit when a held \doPair{} would likely lead to cancellations and (ii) assign larger penalties if the policy unnecessarily delays a \doPair{} that would complete.
For lower tiers, the policy is encouraged to be permissive and avoid holding.

\paragraph{Deployed instantiation.}
The parameterization for base reward shaping is:
\begin{equation}
\eta(a,y) =
\begin{cases}
1 - \kappa_{\text{mis}} \cdot a, & y=\CR,\\
1 + \kappa_{\text{cba}} \cdot a, & y=\CBA,\\
1 + \kappa_{\text{pcaa}} \cdot a, & y=\PCAA,\\
1 + \kappa_{\text{dcaa}} \cdot a, & y=\DCAA,\\
1, & \text{otherwise},
\end{cases}
\label{eq:eta_linear}
\end{equation}
with non-negative coefficients $\kappa_{\text{mis}},\kappa_{\text{cba}},\kappa_{\text{pcaa}},\kappa_{\text{dcaa}}$.
In our tuning, we are using $\kappa_{\text{pcaa}} \ge \kappa_{\text{dcaa}} \ge \kappa_{\text{cba}}$ and $\kappa_{\text{mis}}$ to cap unnecessary holding. Table~\ref{tab:eta_example} shows the design of $\eta(a,y)$ values under a reasonable setting
($\kappa_{\text{mis}}{=}0.15$, $\kappa_{\text{cba}}{=}0.15$, $\kappa_{\text{pcaa}}{=}0.20$, $\kappa_{\text{dcaa}}{=}0.12$), with $a\in\{0,1,2,3\}$.

\begin{table}[t]
  \centering
  \caption{Shaping multipliers $\eta(a,y)$ by tier $a$ and outcome $y$.}
  \label{tab:eta_example}
  \begin{tabular}{lcccc}
    \toprule
    $\eta(a,y)$ & $a=0$ & $a=1$ & $a=2$ & $a=3$ \\
    \midrule
    $y=\CR$   & 1.00 & 0.85 & 0.70 & 0.55 \\
    $y=\CBA$  & 1.00 & 1.10 & 1.20 & 1.30 \\
    $y=\PCAA$ & 1.00 & 1.20 & 1.40 & 1.60 \\
    $y=\DCAA$ & 1.00 & 1.12 & 1.24 & 1.36 \\
    Other     & 1.00 & 1.00 & 1.00 & 1.00 \\
    \bottomrule
  \end{tabular}
\end{table}

\paragraph{Notes.}
This shaping can also be implemented additively (instead of multiplicatively) or with a bounded nonlinear form.
We choose the simple linear form in Eq.~\eqref{eq:eta_linear} because it is easy to audit and tune, and we found it stable in offline iteration and online ablations.

\subsubsection{Temporal shaping}
\label{sec:appendix_time_shaping}

Hold decisions are closely tied to waiting-related experience.
We incorporate a mild bounded temporal term:
\begin{equation}
r_{\text{time}}(s,a) \;=\; \rho_1 \cdot g(\ETA) \;+\; \rho_2 \cdot g(\PWT),
\label{eq:appendix_time}
\end{equation}
where $\ETA$ is estimated pickup time, $\PWT$ is passenger waiting time (or time since call), and $g(\cdot)$ is a clipped monotone transform.
A concrete choice is:
\begin{equation}
g(u)=\text{clip}\Big(\frac{u-u_0}{u_1-u_0},\,0,\,1\Big),
\label{eq:clip_linear}
\end{equation}
with $(u_0,u_1)$ chosen from operational ranges (e.g., $u_0{=}2\text{ min}$, $u_1{=}12\text{ min}$ for time features, and analogous values for distance if used).
In practice, we keep temporal shaping small to avoid overwhelming outcome utility; typical ranges are
$\rho_1,\rho_2\in[0.02,0.10]$.

\paragraph{Final reward.}
The final scalar reward is:
\begin{equation}
r(s,a,y)\;=\; r_{\text{ce}}(s,a,y) \;+\; r_{\text{time}}(s,a).
\label{eq:appendix_final_reward}
\end{equation}
This construction supports experience-aware learning while remaining robust and tunable for deployment.

\subsection{Optimization Derivations}
\label{sec:appendix_opt}

This appendix details Stage~II, which converts experience tiers into hold durations via constrained optimization over empirical quantiles (Section~\ref{sec:opt}).
We provide (i) an equivalent formulation, (ii) solvability notes, and (iii) implementation details used in production.

\subsubsection{Formulation recap and interpretation}
\label{sec:appendix_opt_form}

For each tier $i\in\{0,\dots,K\}$ and outcome category $j\in\mathcal{Y}$, we estimate an empirical CDF:
\begin{equation}
\text{CDF}_{i,j}(x)=\mathbb{P}(T\le x \mid a=i, y=j),
\end{equation}
where $T$ is the system-defined execution clock (e.g., a waiting-related time variable), and we also compute routing fractions $p_{i,j}$.
Given utility weights $v_j$ (consistent with the reward table), we solve:
\begin{align}
\max_{\{x_i\}} \quad &
\sum_{i=0}^{K}\sum_{j\in\mathcal{Y}} p_{i,j}\, v_j\, \text{CDF}_{i,j}(x_i) \label{eq:app_opt_obj}\\
\text{s.t.}\quad &
\sum_{i=1}^{K} p_{i,\CR}\, \text{CDF}_{i,\CR}(x_i)\le \alpha, \label{eq:app_opt_con}\\
& 0=x_0\le x_1\le \cdots \le x_K \le x_{\max}. \label{eq:app_opt_mono}
\end{align}

For a fixed tier $i$, $\text{CDF}_{i,j}(x_i)$ is the fraction of outcome-$j$ events that would be ``covered'' by holding for duration $x_i$ under the execution clock $T$.
The objective aggregates expected utility across outcomes; the constraint bounds the fraction of completion cases that are unnecessarily held, controlled by $\alpha$.

\subsubsection{Solvability and structure}
\label{sec:appendix_opt_solve}

\begin{table}[t]
  \centering
  \caption{Stage~II optimization parameters and ranges.}
  \label{tab:stage2_params}
  \begin{tabular}{lcc}
    \toprule
    Parameter & Applied value\\
    \midrule
    Tiers $K$ (actions $|\mathcal{A}|{=}K{+}1$) & 3 \\
    Guardrail $\alpha$ (completion mis-hold cap) & 0.01 \\
    Max hold $x_{\max}$ & 30s \\
    Quantile grid step & 0.01 \\
    Recalibration cadence & weekly \\
    Data window for CDF/quantiles & 7 days \\
    \bottomrule
  \end{tabular}
\end{table}

Although $\text{CDF}_{i,j}(\cdot)$ is non-decreasing, the optimization is not necessarily convex because empirical CDFs are step functions.
However, the problem is low-dimensional ($K$ is small in our production) and admits efficient exact or near-exact solutions. Because each $\text{CDF}_{i,j}(x)$ changes only at observed sample values of $T$, the optimizer only needs to consider a discrete candidate set $\mathcal{G}_i$ for each tier $i$.
Our practical choice is to use quantile grid points:
\begin{equation}
\mathcal{G}_i = \{Q_i(q)\mid q \in \{0, 0.01, 0.02,\dots,0.99\}\},
\end{equation}
where $Q_i(q)$ is the empirical quantile of $T$ among tier-$i$ samples.
This converts the problem into a finite constrained search.

\paragraph{DP under monotonicity.}
Under the monotone constraint $x_0\le x_1\le \cdots \le x_K$, we can solve the discretized problem by dynamic programming.
Let $\mathcal{G}_i$ be sorted in ascending order and define
\begin{equation}
U_i(x) = \sum_{j\in\mathcal{Y}} p_{i,j}\, v_j\, \text{CDF}_{i,j}(x),
\qquad
C_i(x) = p_{i,\CR}\, \text{CDF}_{i,\CR}(x).
\end{equation}
Then the problem becomes:
\begin{equation}
\max_{x_i \in \mathcal{G}_i} \sum_{i=0}^{K} U_i(x_i)
\quad \text{s.t.}\quad
\sum_{i=1}^{K} C_i(x_i) \le \alpha,\ \ x_0 \le x_1 \le \cdots \le x_K.
\label{eq:dp_form}
\end{equation}
With a discretized constraint budget, the DP complexity is modest.
In practice, $K{=}3$ and $|\mathcal{G}_i|$ is on the order of 50--200, making the solver easily fast enough for frequent recalibration. Table~\ref{tab:stage2_params} summarizes the parameters used in Stage~II optimization.

\section{Alternating Training for Stable Tiering under Simulator Imperfections}
\label{sec:appendix_alt_training}

Here we provide additional details on the alternating training procedure used in Stage~I of \method.
The goal is to obtain a tiering policy that is (i) stable under representation drift, (ii) robust to non-stationary traffic regimes, and (iii) safe to deploy in DiDi's production matching system where rapid rollback and predictable behavior are required.

\subsection{Why Alternating Updates are Necessary}
\label{sec:appendix_alt_motivation}

Stage~I combines an encoder $f_{\theta}$ with a contextual bandit head parameterized by $\phi$.
Although one could jointly update $(\theta,\phi)$, we found such training unstable in deployment-oriented settings.
The main reason is that the bandit head relies on embeddings $z_t=f_{\theta}(h_t)$ to maintain calibrated statistics.
When the encoder changes continuously, the feature space drifts, which can invalidate the accumulated bandit statistics and lead to abrupt tier-distribution shifts.

A second issue is simulator-induced bias.
Exploration is necessary for improving the tiering policy beyond logged behavior, but aggressive exploration in a simulator may exploit artifacts that do not transfer to production.
If the encoder and bandit are updated simultaneously under such biased trajectories, the entire policy can drift toward unstable or non-transferable behavior.

Therefore, we use alternating updates to decouple representation learning from bandit policy improvement, which yields smoother tier evolution and more predictable rollout behavior.

\subsection{Training Data Sources and Notation}
\label{sec:appendix_alt_data}

We assume access to an offline log dataset $\mathcal{D}_0$ collected under a strong production policy and an interaction environment $\mathcal{E}$ for iterative improvement.
In our implementation, $\mathcal{E}$ is instantiated as a simulator, though the same procedure can also use controlled traffic slices under stricter deployment constraints.
Each interaction produces a tuple $(h_t, a_t, r_t, y_t)$, where $h_t$ is the sequential context, $a_t$ is the selected tier, $r_t$ is the experience reward defined in Section~\ref{sec:reward_design}, and $y_t$ is the realized outcome category.

The encoder maps the context to an embedding $z_t=f_{\theta}(h_t)$, and the bandit head selects tiers according to the UCB rule in Eq.~\eqref{eq:ucb}.
For encoder training, we define a utility label $\tilde{R}_t$ as a short-horizon discounted return:
\begin{equation}
\tilde{R}_t = \sum_{\tau=0}^{H-1}\gamma^{\tau} r_{t+\tau},
\label{eq:appendix_return}
\end{equation}
where $H$ is the horizon length and $\gamma\in(0,1]$ is the discount factor.

\subsection{Alternating Training Procedure}
\label{sec:appendix_alt_procedure}

We adopt alternating optimization with explicit freezes to decouple representation learning from exploration.
Each outer iteration consists of two phases:

\paragraph{Phase A: Bandit adaptation with frozen encoder.}
We freeze $\theta$ and update the bandit head $\phi$ by interacting with the simulated environment $\mathcal{E}$.
Because the representation is fixed, the bandit statistics remain well-defined and the tier distribution evolves smoothly.
This phase is responsible for controlled exploration and fast adaptation of tier selection.

\begin{algorithm}[t]
\caption{Alternating Training for Stable Tiering}
\label{alg:appendix_alt_train}
\begin{algorithmic}[1]
\Require Offline logs $\mathcal{D}_0$, simulated environment $\mathcal{E}$, outer iterations $K$,
bandit steps $T_{\text{bandit}}$, encoder steps $T_{\text{enc}}$
\State Initialize encoder parameters $\theta$ (supervised pretraining on $\mathcal{D}_0$)
\State Initialize bandit parameters $\phi$
\For{$k=1$ to $K$}
  \State \textbf{Phase A (Bandit):} freeze $\theta$
  \For{$t=1$ to $T_{\text{bandit}}$}
    \State Observe $h_t$, compute $z_t=f_{\theta}(h_t)$, select $a_t$ via Eq.~\eqref{eq:ucb}
    \State Execute tier $a_t$ in $\mathcal{E}$
    \State Observe $(r_t,y_t)$, append $(h_t,a_t,r_t,y_t)$ to $\mathcal{D}_k$
    \State Update $\phi$ using $(z_t,a_t,r_t)$
  \EndFor
  \State \textbf{Phase B (Encoder):} freeze $\phi$
  \State Construct labels $\tilde{R}_t$ on $\mathcal{D}_0 \cup \mathcal{D}_{1:k}$ (Eq.~\eqref{eq:appendix_return})
  \For{$u=1$ to $T_{\text{enc}}$}
    \State Sample batches from $\mathcal{D}_0 \cup \mathcal{D}_{1:k}$ and update $\theta$ using Eq.~\eqref{eq:appendix_repr_obj}
  \EndFor
\EndFor
\State \Return Tiering policy $\pi_{\theta,\phi}$
\end{algorithmic}
\end{algorithm}

\paragraph{Phase B: Representation improvement with frozen bandit.}
We freeze the bandit head and update the encoder on aggregated trajectories (offline logs plus interaction data).
Because the decision rule is fixed during this phase, the encoder learns to better predict experience-aligned utility labels without introducing additional exploration-induced non-stationarity.

Concretely, given the aggregated dataset $\mathcal{D}_{\le k}=\mathcal{D}_0\cup\mathcal{D}_1\cup\cdots\cup\mathcal{D}_k$, we optimize the encoder objective:
\begin{equation}
\min_{\theta}\ \mathbb{E}_{(h_t,\tilde{R}_t)\sim \mathcal{D}_{\le k}}\big[\ell(g_{\theta}(h_t),\tilde{R}_t)\big],
\label{eq:appendix_repr_obj}
\end{equation}
where $\ell$ is a robust regression loss (e.g., Huber) and $g_{\theta}$ is a prediction head on top of the Transformer encoder. The overall training procedure is shown in Algorithm~\ref{alg:appendix_alt_train}.

\end{document}